\documentclass[12pt, draftclsnofoot, onecolumn]{IEEEtran}

\usepackage{ifpdf}
\usepackage{cite}
\usepackage[pdftex]{graphicx}

\usepackage[T1]{fontenc}
\usepackage{enumitem}
\usepackage{afterpage}
\usepackage{bm}
\usepackage{listings}
\usepackage{xcolor}
\usepackage{graphicx}
\usepackage{float}
\usepackage[font=small]{caption}
\usepackage[font=footnotesize]{subcaption}

\usepackage{placeins}
\usepackage{url}
\usepackage{epstopdf}
\usepackage{booktabs}
\usepackage{footnote}
\makesavenoteenv{table}
\makesavenoteenv{tabular}
\usepackage{flushend}
\usepackage{amsmath}
\usepackage{amssymb}
\usepackage{tabularx}
\usepackage{booktabs}
\usepackage{placeins}
\usepackage{graphicx}
\usepackage{color}
\usepackage{colortbl}
\usepackage{bbold}

\usepackage[acronyms,nonumberlist,nopostdot,nomain,nogroupskip]{glossaries}

\makeatother

\newcommand{\edit}[1]{\textcolor{black}{#1}}

\usepackage{glossaries}
\newacronym{iot}{IoT}{Internet of Things}
\newacronym{ml}{ML}{Machine Learning}
\newacronym{marl}{MARL}{Multi-Agent Reinforcement Learning}
\newacronym{rl}{RL}{Reinforcement Learning}
\newacronym{decpomdp}{Dec-POMDP}{Decentralized Partially Observable Markov Decision Process }
\newacronym{pomdp}{POMDP}{Partially Observable Markov Decision Process}
\newacronym{uav}{UAV}{Unmanned Aerial Vehicle}
\newacronym{dqn}{DQN}{Deep Q-Network}
\newacronym{dnn}{DNN}{Deep Neural Network}
\newacronym{dial}{DIAL}{Differentiable Inter-Agent Learning}
\newacronym{mdp}{MDP}{Markov Decision Process}
\newacronym{fov}{FoV}{Field of View}
\newacronym{cnn}{CNN}{Convolutional Neural Network}
\newacronym{nn}{NN}{Neural Network}
\newacronym{ddql}{DDQL}{Distributed Deep Q-Learning}
\newacronym{pdf}{PDF}{Probability Density Function}
\newacronym{ndpomdp}{ND-POMDP}{Networked Distributed Partially Observable Markov Decision Process}
\newacronym{radam}{RAdam}{Rectified Adam}
\newacronym{cdf}{CDF}{Cumulative Distribution Function}
\newacronym{mpc}{MPC}{Model Predictive Control}
\newacronym{rv}{rv}{Random Variable}
\newacronym{qoe}{QoE}{Quality of Experience}
\begin{document}

\title{Distributed Reinforcement Learning for Flexible and Efficient UAV Swarm Control}

\author{
\IEEEauthorblockN{\large Federico Venturini$^*$, Federico Mason$^*$, Francesco Pase$^*$,\\Federico Chiariotti$^{\sharp}$, Alberto Testolin$^{*\dagger}$, Andrea Zanella$^*$, Michele Zorzi$^*$ \vspace{1mm}}
    
    \IEEEauthorblockA{
		\small $^*$Department of Information Engineering, University of Padova - Via Gradenigo, 6/b, Padova, Italy
	}
	
	\IEEEauthorblockA{
		\small $^\sharp$ Department of Electronic Systems, Aalborg University - Fredrik Bajers Vej 7C, Aalborg, Denmark
	}
	
	\IEEEauthorblockA{
		\small $^\dagger$Department of General Psychology, University of Padova - Via Venezia, 8, Padova, Italy
	}

	\IEEEauthorblockA{
		\small $^{*\dagger}$\small\texttt{\{venturini, masonfed, pasefran, testolin, zanella, zorzi\}@dei.unipd.it},
		\small $^\sharp$\small\texttt{fchi@es.aau.dk}
    }
}

\maketitle

\begin{abstract}
Over the past few years, the use of swarms of \glspl{uav} in monitoring and remote area surveillance applications has become widespread thanks to the price reduction and the increased capabilities of drones.
The drones in the swarm need to cooperatively explore an unknown area, in order to identify and monitor interesting targets, while minimizing their movements.
In this work, we propose a distributed Reinforcement Learning (RL) approach that scales to larger swarms without modifications.
The proposed framework \edit{relies on the possibility for the \glspl{uav} to exchange some information through a communication channel, in order to achieve context-awareness and implicitly coordinate the swarm's actions.}
Our experiments show that \edit{the proposed method can yield effective strategies, which are robust to communication channel impairments, and that can easily deal with non-uniform distributions of targets and obstacles. Moreover,} when agents are trained in a specific scenario, they can adapt to a new one with \edit{minimal} additional training.
We also show that our approach achieves better performance compared to a computationally intensive look-ahead heuristic.
\end{abstract}

\glsresetall


\IEEEpeerreviewmaketitle

\section{Introduction}\label{sec:intro}
\glsresetall

The high data rate achievable with modern wireless communications and the increasing computational power of embedded systems, along with the sharp price reduction of commercial \glspl{uav}, have enabled the use of swarms of drones for Smart City services~\cite{HosseinMotlagh2016}.
Thanks to their size, flexibility and flight ability, these swarms represent a new solution for a plethora of different applications, such as remote surveillance, distributed sensing, wireless coverage extension and object tracking~\cite{Shakhatreh2019}.

Over the past few years, researchers have proposed several \gls{uav}-based systems~\cite{Shakeri2019}, but achieving an efficient distributed control is a complex problem, whose solution is often task-dependent.
In this context, it is important to properly define the different sub-tasks of surveillance, monitoring, mapping and tracking \cite{Chung2018}.
In this work, we assume that targets are static, but occupy random positions in the monitored area.
Moving \glspl{uav} are equipped with sensors that can detect targets within a limited sensing range, \edit{and a radio interface that makes it possible to share position information and sensing data. The UAVs need to coordinate} to explore the area and find the targets \edit{without colliding with each other or with obstacles}.   

The problem of identifying fixed targets arises in several practical situations, ranging from the generation of real-time flood maps~\cite{Baldazo2019} to the detailed tracking of weeds in agriculture~\cite{Albani2017}, but an efficient initial exploration is of interest even \edit{for} larger classes of problems, e.g., considering moving targets.
\edit{One such example is wildfire monitoring in dry regions~\cite{Julian2019}, which can be effective as long as the \glspl{uav} move faster than the spread of the fires.}

The dynamic nature of these problems, in which actions can have long-term consequences and affect the future evolution of the environment in complex ways, makes them a natural application area for \gls{rl} techniques~\cite{sutton1998introduction}. However, due to the curse of dimensionality, \edit{a centralized approach to the problem} (i.e., using a single controller) is feasible only for very small swarms.
In order to design a scalable system, \gls{marl} techniques need to be used, but the non-stationarity of the environment~\cite{Hernandez-Leal2017} complicates the system design and the agent training.
This additional complexity makes \gls{marl} an open research field, and the different degrees of centralization and communication between agents make the configuration of the learning system \edit{an interesting problem to investigate}. 

In this work, we consider a \gls{marl} framework for exploration and surveillance.
Our aim is to find a flexible \gls{ml} strategy to explore and monitor a certain area with a swarm of \glspl{uav} \edit{that can exchange information within a certain coverage range.}
Performance is determined by the ability of the drones to find and reach the targets, which are located in unknown positions.

\edit{In our framework, the} observations of other agents are \edit{shared through a radio channel and} used to make decisions and to avoid collisions, thus encouraging cooperation. We define a \gls{dqn} algorithm and demonstrate its efficiency with limited training, comparing it to a benchmark look-ahead heuristic and showing that our approach can better explore the environment and reach the targets faster.
We also perform a transfer learning experiment, showing that agents trained on a \edit{certain} map can learn to adapt to a completely new scenario much faster than restarting the training from scratch.

\edit{We adopted a general model, using a grid-world representation and making a limited number of assumptions on the nature of the task. Nonetheless, we show that our system can be implemented in several different scenarios.
In particular, the map is not entirely visible to the \glspl{uav}, there are obstacles, and targets are in unknown positions (often clustered together, making clusters rarer and thus harder to find).
These features make \gls{marl} highly complex, especially when considering limited communication capabilities: to the best of our knowledge, our work is the first to apply it in such challenging conditions.}

\edit{Our approach to solve the problem is to model the state as a series of correlated maps, which contain different information on the environment, making the learning framework extendable to even more complicated scenarios.}

\edit{
The contributions of this paper can be summarized as follows:
\begin{itemize}
 \item We formulate a \gls{ndpomdp} framework for swarm management in a complex environment and propose a \gls{marl} architecture to address such a problem;
 \item We show that the proposed system can outperform computationally heavy heuristics and transfer its knowledge to different scenarios with limited retraining;
 \item We analyze the effect of bigger changes in the environment, such as changing the size of the map or the number of drones, and show that transfer learning is still effective;
 \item We show that the system is robust to channel impairments, and can perform very well even in realistic scenarios that differ from the more abstract models used in the training phase.
\end{itemize}
}

A preliminary version of this paper was presented at ACM DroNet 2020~\cite{venturini2020distributed}; this version has a significantly updated system model, considering different map sizes and the presence of obstacles as well as a different \gls{marl} solution, and more extensive results on the performance of our approach. \edit{Moreover, we have added the analysis of the impact of the communication channel on the system's performance, \edit{and tested} the proposed solution in a map obtained from real data.}

The rest of the paper is divided as follows: first, Sec.~\ref{sec:related} analyzes the related work in the field. The system model and \gls{marl} algorithm are presented in Sec.~\ref{sec:model}. The experimental setup is reported in Sec.~\ref{sec:setup}, while the experimental results, including transfer learning experiments, are reported in Sec.~\ref{sec:results}. Finally, Sec.~\ref{sec:conclusions} concludes the paper and presents some possible avenues of future work.

\section{Related Work}\label{sec:related}
An extensive taxonomy of multi-agent solutions was presented in \cite{Busoniu2010}. The general approaches adopted to solve the \gls{marl} problem can be cast into one of these four frameworks: $(1)$ a single agent architecture that interacts with multiple copies of itself, generating emergent behaviors; $(2)$ communication between agents of the same type with improved coordination; $(3)$ cooperation between agents with different specialized goals achieving coordinated behavior; and $(4)$ modeling other agents' behaviors and planning a response \cite{Hernandez-Leal2019}.

The authors in \cite{Zanol2019} study the first of these four approaches and use the tabular Q-learning algorithm to guide drones to survey an unknown area, showing that even the simplest \gls{marl} algorithm can improve the overall system rewards. Similarly, in \cite{Cui2019} and \cite{Cui2020} the \gls{marl} framework is applied to a more complex problem in which a \gls{uav} network is adopted to provide flexible wireless communication. However, in these works the \gls{marl} algorithm is used to optimize resource allocation instead of guiding drones, so that a coordinated exploration strategy is missing. 

An interesting research direction \edit{for} \gls{marl} is pioneered in \cite{Foerster2016}, which uses \glspl{dnn} to represent and learn more complex Q-functions \cite{Mnih2015}. At first, the authors study the performance of one network trained for all agents, which then share the same parameters during the execution phase (this is also our approach). A second proposed system uses the \gls{dial} framework, in which agents learn meaningful real-valued messages to be exchanged in order to improve cooperation: this allows for faster training, but the model is limited to a very small number of agents.

Other works use \gls{rl} in the practical scenarios discussed above:
in~\cite{Baldazo2019}, the authors adopt a \gls{marl} approach to control a flood-finding swarm of \glspl{uav}. However, the model only considers a swarm with a fixed number of drones, and the experimental results are not compared to state-of-the-art heuristics.
In \cite{Albani2017}, a reinforced random walk model is exploited to map weeds in an agricultural setting, taking noisy acquisition into account and solving the issue with collective observations. Random walks are then biased based on the positions of the already discovered targets, which have to be properly mapped, along with the distances from other drones in the network. In this case, the authors considered swarms of variable sizes, but the random walk needs to be manually tuned for each setting.
Another recent study~\cite{Julian2019} considers wildfire spread monitoring, checking how the fire evolves and spreads in the map \edit{from a known starting point}. The authors define the problem as a \gls{decpomdp}~\cite{oliehoek2016concise} and carry out several experiments, as well as comparisons against a greedy heuristic (similar to the look-head method we studied in this work).
A target-tracking application for disaster scenarios, with a model similar to our own but applied to a single drone, is described in~\cite{wu2019uav}.
\edit{
Finally,~\cite{Albani2017} considers a \gls{marl} system with realistic communication, where a swarm of drones needs to explore a map, receiving a reward for each new explored location.
This is a simpler reinforcement \edit{learning} problem, as identifying and finding rare targets is much more challenging to learn due to the delayed reward.}

The \gls{marl} approaches can also fit models in which \gls{uav} connectivity is important: in \cite{challita2018deep}, a framework including \gls{rl} and game theory is used to plan the path of two drones that need to save energy and minimize the interference to the ground network while maintaining a cellular connection.
Furthermore, in~\cite{liu2018energy} the authors design a centralized \gls{rl} system to maximize coverage for a swarm of aerial base stations serving mobile users on the ground.
A similar approach is taken in~\cite{liu2019reinforcement}, which redefines the problem in terms of \gls{qoe} maximization for the users.
For a fuller communication-oriented perspective on the use of \gls{rl} for \gls{uav} networks, we refer the reader to~\cite{hu2020reinforcement}.

\edit{These works have similar objectives to our own, but either go back to the single-agent setting or have restrictive assumptions: as an example,~\cite{Julian2019} considers well-known fire patterns, which can be extensively learned, with a known starting point.
In our case, the initial positions of the targets and \edit{ of the \glspl{uav} are} not the same across different episodes, making the model more general and complicating the learning task.
Furthermore, unlike previous efforts in the literature, we exploit the transfer learning paradigm, showing how our model can easily adapt to scenarios with obstacles, realistic maps, and different swarm sizes. To the best of our knowledge, our work presents the most complex environment to date, in which a single architecture can deal with different map and swarm sizes, different numbers of targets to track, and the presence of obstacles.}

\section{System Model} \label{sec:model}

In the following, we first present the environment \edit{where the \glspl{uav} operate}. We give a full list of the notation used in Table~\ref{tab:sim_notation} as a reference to the reader.

\begin{table}[t]\centering
    \footnotesize
	\begin{tabular}[c]{cl|cl}
		\toprule
		Symbol & Description & Symbol & Description \\
		\midrule
		$\mathcal{M}$ & Coordinate set & $\mathcal{O}$& Observation space of the system \gls{ndpomdp}\\
		$M$ & Map grid size & $\bm{\Phi}$ & Matrix of cell values\\
		$K$ & Number of targets &  $\mathbf{X}$ & Matrix of \gls{uav} positions\\
		$\mathbf{z}_k$ & Coordinates of the $k$-th target &  $\bm{\Omega}$ & Matrix of obstacle positions \\
		$\sigma$ & Standard dev. of the target Gaussian functions & $\hat{\bm{\Phi}}$& Observed cell value matrix\\
		$\phi(\cdot)$ & Cell value function &$\hat{\bm{\Omega}}$& Observed obstacle position matrix\\
		$\mathcal{U}$ & Set of \glspl{uav} & $\mathbf{X}_u$ & Observed \gls{uav} position matrix for $u$\\ 
		$U$ & Number of \glspl{uav} \edit{(cardinality of $\mathcal{U}$)} & $F$& Observation window size \edit{(in number of cells)} \\ 
		$d_{\text{sparse}}$ & Minimum target distance in the sparse scenario &  $\psi$ & Penalty for collisions\\
		$\omega(\cdot)$ & Obstacle location function & $\theta$ & Penalty for moving to forbidden areas\\
		$\eta$ & Fraction of the map occupied by obstacles & $\rho$ & Obstacle value \\
		$\zeta$ & \acrlong{fov} of each \gls{uav} & $\nu_u(\mathbf{x}_u,\mathbf{a}_u)$ & Invalid move indicator function for \gls{uav} $u$\\
		$h_{\min}$ & Minimum obstacle size \edit{(in number of cells)} & $\chi_u(\mathbf{X},\mathbf{A})$ & Collision indicator function for \gls{uav} $u$\\
		$h_{\max}$ & Maximum obstacle size \edit{(in number of cells)} & $r_u(s,\mathbf{a})$ & Reward for \gls{uav} $u$ \\
		$\bm{\ell}_i$ & Lower left corner coordinates of the $i$-th obstacle & $\pi$ & Observation-action policy\\
		$\mathcal{H}_i$ & Set of cells occupied by the $i$-th obstacle &$R_{u,t}(\pi)$ & Long-term reward for $u$ using policy $\pi$ \\
		$\mathbf{h}_i$ & \edit{Size} of the $i$-th obstacle &  $\gamma$ & Exponential discount factor\\
		$N$ & Episode duration \edit{(steps)} & $e_t$ & Experience sample\\
		$\mathcal{S}$ & State space of the system \gls{ndpomdp} &$\alpha$ & Learning rate \\
		$\mathcal{V}(\mathbf{s})$ & Valid move space for state $s$ &   $B_{\text{size}}$ & Size of a learning batch\\
        $\mathcal{A}$ & Action set &$Q(o_u,a_u)$ & Q-value estimate of $R$\\
        $\mathbf{a}_u$ & Action for \gls{uav} $u$ & $n_q$ & Model update period \edit{(steps)} \\
		\bottomrule
	\end{tabular}
	\vspace{0.1cm}
	\caption{Notation definitions.}
	\label{tab:sim_notation}\vspace{-0.8cm}
\end{table}

\subsection{Environment}

The system environment consists of a square grid of size $M \times M$.
Each cell of the grid \edit{(we will refer to a cell or a location interchangeably in the following)} is identified by its coordinates $\mathbf{m} \in \mathcal{M}$, where \edit{$\mathcal{M} = \mathcal{X} \times \mathcal{Y}$, and $\mathcal{X}=\mathcal{Y}=\{0,...,M-1\}$}. We place a set of $K$ targets on the map, which represent the objectives of the \gls{uav} surveillance application. The position of the $k$-th target is denoted as $\mathbf{z}_k=(x_k,y_k)$.

We then generate a set of $K$ bivariate Gaussian functions over the grid, which represent the visibility of a target to the \glspl{uav}, with the same covariance matrix $\Sigma = \bigl( \begin{smallmatrix}\sigma^2 & 0\\ 0 & \sigma^2\end{smallmatrix}\bigr)$.
The mean $\mathbf{z}_k=(z_{k,1},z_{k,2})$ corresponds to the coordinates of the target. 
Note that the Gaussian functions do not represent actual distributions, but rather the full view of the \glspl{uav}, which can see a target from afar. The value of $\sigma$ can be interpreted as the distance at which a target can be identified, as larger values of $\sigma$ mean that the target is visible from further away.

Each cell can then be associated with a weight $\phi(\mathbf{m})$, which represents the \textit{value} of the location, which increases with the proximity to a target, and is given by the maximum of the Gaussian functions in that point, normalized in such a way that the target locations have values equal to $1$:
\begin{equation}
\label{eq:map_value}
    \phi(\mathbf{m}) = \max_{k\in\{0,\ldots,K-1\}} e^{- \frac{1}{2} \left((\mathbf{m}-\mathbf{z}_k)^\mathbf{T}\Sigma(\mathbf{m}-\mathbf{z}_k)\right)}.
\end{equation}
If $\phi(\mathbf{m})$ is smaller than 0.01, it is set to 0, as the \glspl{uav} cannot see \edit{any} target from that location.
Under these conditions, the most valuable cells coincide with the center of each Gaussian function, which represents one of the targets in the considered scenario.
While the environment is static, the \glspl{uav} move within the map with the aim of positioning themselves over the targets as fast as possible. We denote the set of \glspl{uav} by $\mathcal{U}$, and by $U$ its cardinality.

\begin{figure}[t!]
\centering
\includegraphics[width=.7\textwidth]{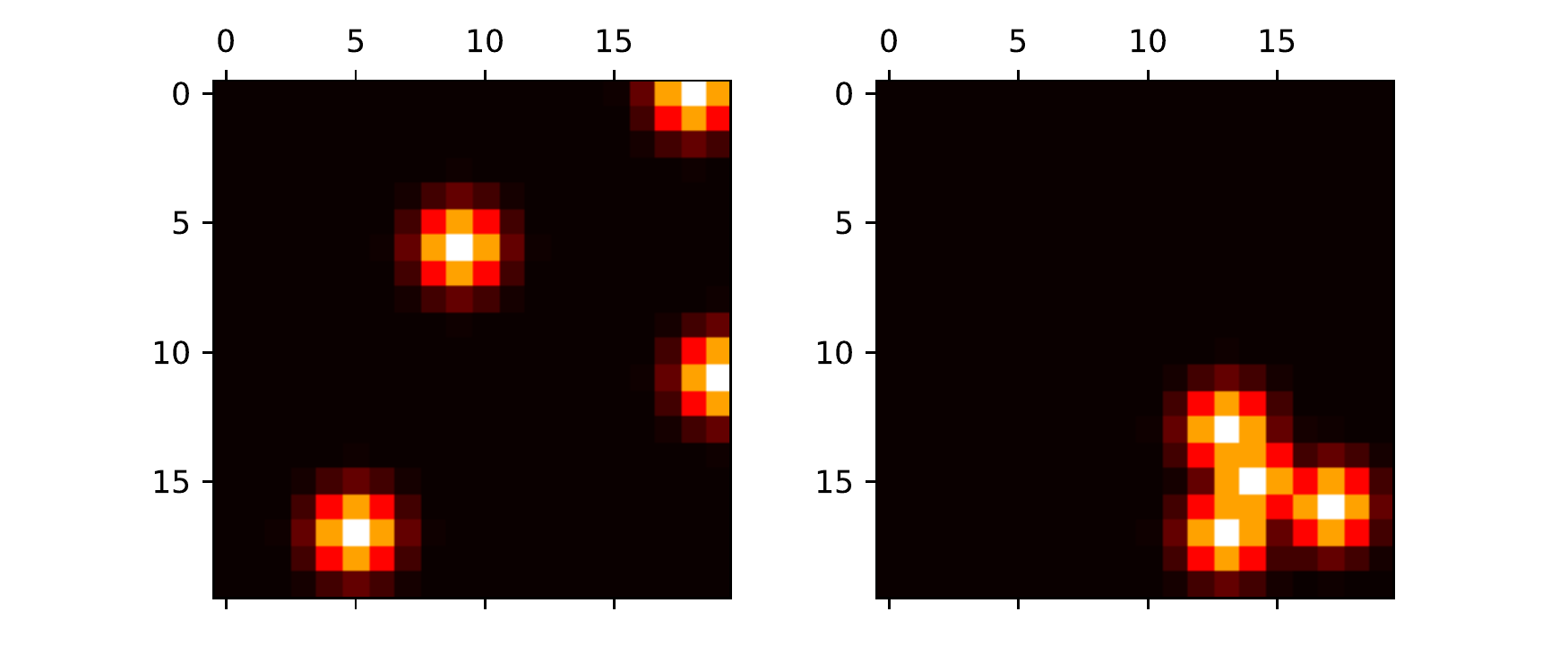}
\caption{Two examples of the sparse (left) and cluster (right) target distributions.\vspace{-0.3cm}}
\label{fig:distributions}
\end{figure}

In this work, we consider two different distributions for the targets, named \emph{sparse} and \emph{cluster}, which are characterized by different correlations among the target positions. In both cases, the first target is randomly placed on the grid following a 2D uniform distribution: $\mathbf{z}_0$ can take any value in $\mathcal{M}$ with equal probability. The other targets are then placed sequentially, according to the following rules. 
In the sparse scenario, the position $\mathbf{z}_i$ of the $i$-th target is randomly chosen in the set $\mathcal{M}^{\text{sparse}}_i = \{\mathbf{m} \in \mathcal{M}: ||\mathbf{m}-\mathbf{z}_j||_2 > d_{\text{sparse}} ,\: \forall j < i \}$, with probability mass distribution
\begin{equation}
\label{eq:target_sparse}
    P_{\text{sparse}}(\mathbf{z}_i = \mathbf{m}) =
    \frac{||\mathbf{m}-\mathbf{z}_0||_2}{\kappa_i^\text{sparse}}.
\end{equation}
where \edit{$\kappa_i^\text{sparse}=\sum_{\mathbf{m} \in \mathcal{M}^{\text{sparse}}_i}  ||\mathbf{m}-\mathbf{z}_0||_2$} is a normalization factor. Hence, the other targets tend to be distributed far from the first, with  a minimum distance $d_{\text{sparse}}$ between each other.

In the cluster scenario, instead, the $i$-th target can take any position in the set $\mathcal{M}^{\text{cluster}}_i = \{\mathbf{m} \in \mathcal{M}: ||\mathbf{m}-\mathbf{z}_j||_2 > 1, \: \forall j < i \}$ with probability mass distribution
\begin{equation}
\label{eq:target_cluster}
P_{\text{cluster}}(\mathbf{z}_i = \mathbf{m}) =
    \frac{1}{(1+||\mathbf{m}-\mathbf{z}_0||_2)\kappa_i^\text{cluster}}.
\end{equation}
where \edit{$\kappa_i^\text{cluster}=\sum_{\mathbf{m} \in \mathcal{M}^{\text{cluster}}_i} \frac{1}{(1+||\mathbf{m}-\mathbf{z}_0||_2)}$} is a normalization factor. In this case, the targets tend to cluster around the first one, but cannot occupy adjacent cells, since the minimum distance must be greater than 1. 

An example of the two target placements is shown in Fig.~\ref{fig:distributions}.
These two distributions represent two plausible configurations of targets in tracking applications: in wildlife monitoring, some species of animals might tend to herd together, while more territorial ones will have a sparser distribution on the map.
The same goes for a battlefield scenario, in which groups of soldiers might act together as a tight formation, while guerrilla-style fighting will involve a much sparser distribution of forces.

In a more complex version of the scenario, the map does not just have targets that the \glspl{uav} need to find and reach, but \emph{obstacles} as well: in an urban scenario, these might be tall buildings or designated no-fly zones, while in a natural scenario they might correspond to natural obstacles such as \edit{boulders} or tall trees. We define a function $\omega(\mathbf{m})$, which is equal to 1 if the cell corresponds to an obstacle and 0 otherwise. Then, we denote by $\eta$ the portion of the map occupied by obstacles:
\begin{equation}
  \eta=\sum_{\mathbf{m}\in\mathcal{M}}\frac{\omega(\mathbf{m})}{M^2}.
\end{equation}
\edit{Cells inside an obstacle are considered impassable, like the map borders, and the \glspl{uav} that try to move on an obstacle will remain in the same cell.}

\edit{For the training of our algorithm, we assumed that obstacles are rectangular and randomly scattered in the area.} The $i$-th obstacle is determined by its dimensions $\mathbf{h}_i$ and by the position of its lower left corner $\bm{\ell}_i$. We formally define the obstacle as the set $\mathcal{H}_i$:
\begin{equation}
  \mathcal{H}_i=\left\{\mathbf{m}=(m_1,m_2)\in\mathcal{M}:m_1\in\{\ell_{i,1},\ldots,\ell_{i,1}+h_{i,1} \edit{-1} \},m_2\in\{\ell_{i,2},\ldots,\ell_{i,2}+h_{i,2} \edit{-1}\}\right\}.
\end{equation}

Obstacles are generated sequentially, like the targets, and for each obstacle $i$ the dimensions  $\mathbf{h}_i$ are drawn uniformly from the set \edit{$\{h_{\min},h_{\max}\} \times \{h_{\min},h_{\max}\}$}. The lower left corner position $\bm{\ell}_i$ is then drawn from a uniform distribution in the set $\mathcal{M}^{\text{obs}}_i$, the subset of the map defined as:
\begin{equation}
  \mathcal{M}^{\text{obs}}_i\!=\!\{\bm{\ell}\in\mathcal{M}: \mathcal{H}_i\subset\mathcal{M},||\mathbf{n},\mathbf{z}_k||_2>1,\forall \mathbf{n}\in\mathcal{H}_i,k\in\{0,\ldots,K-1\}, d(\mathcal{H}_i,\mathcal{H}_j)\geq2,\forall j<i\},
\end{equation}
where $d(\mathcal{H}_i,\mathcal{H}_j)=\min_{\mathbf{m}_i\in\mathcal{H}_i,\mathbf{m}_j\in\mathcal{H}_j}||\mathbf{m}_i-\mathbf{m}_j||_2$ is the distance between the obstacles $i$ and $j$.
The three constraints force the obstacle to be entirely inside the map, not to be directly adjacent to any of the targets, and not to touch other obstacles. The choice of these constraints was motivated by the necessity to \edit{guarantee the existence of a clear} path to the targets from any point in the map.

We consider multiple \emph{episodes} of $N$ steps: in each episode, the targets, \glspl{uav}, and obstacles are redistributed in the map, and the swarm must locate the targets in as few steps as possible. We consider discrete time slots, so that each drone can move by a single cell at each time step. Furthermore, we assume that a \gls{uav} has a limited \gls{fov}, i.e., it can only know the value of the cells within a radius $\zeta$.
This framework allows us to represent many different applications and scenarios by changing the size of the grid, the number of drones, targets and obstacles, the \gls{fov} range $\zeta$ and the target visibility parameter $\sigma$. It can also be easily extended to dynamic targets.

At the beginning of each episode, each \gls{uav} only knows the values of the cells within the swarm's \gls{fov}. The drones assume that all unexplored points of the map are associated with the maximum $\phi(\mathbf{m})$. Then, each \gls{uav} moves independently at each time step $n$: as the swarm explores the environment, each drone discovers the values of the map locations that it has covered, and updates its information according to $\phi(\mathbf{m})$. We highlight that the knowledge about the map is instantly shared, which means that each drone receives the observations that all the other drones have acquired. \edit{This is always true during training, whereas in some testing episodes we also experiment the scenarios in which unreliable communications affect the shared messages.} The objective of the swarm for each episode is to position each of its \glspl{uav} above a target as quickly as possible.

\edit{\subsection{Communication model}}
\label{sec:comm_model}

We consider the swarm to only have partial observations: as the size of the map might be too large for the swarm to effectively coordinate over it, we consider each \gls{uav} to have up-to-date knowledge only inside the $F\times F$ square with it at the center, with $F\leq M$. If the distance between the \gls{uav} and the edge of the map is lower than $F$, \edit{the square will consider the edge of the map as the edge of the visible region, and the \gls{uav} will no longer be at its center, in order to avoid modeling the area outside the map}. This assumption allows us to model communication constraints in the problem, as \glspl{uav} need to share the observed parts of the map with the other components of the swarm; however, $F$ should not be confused with the \gls{fov} $\zeta$, as the former represents the size of the portion of the map that each \gls{uav} considers when deciding its next action, while the latter represents the size of the portion of the map that the \gls{uav} can sense directly at each moment. \edit{In our case, we always have $F > \zeta$.}

\subsection{ND-POMDP formulation}
\label{sec:pomdp}

The described scenario is modeled as an \gls{ndpomdp}~\cite{nair2005networked}, i.e., a \gls{mdp} where the system state in not directly observable and is influenced by the actions of multiple agents, whose behavior is not centrally coordinated. Indeed, the swarm only has limited knowledge of the map, and the \glspl{uav} can take actions independently and have independent rewards. \edit{We observe that \gls{ndpomdp} is a particular class of Decentralized POMPD (\gls{decpomdp}) for which not all agents interact with each other~\cite{kumar2011scalable}. Convergence to the optimal solution for this kind of problem has been proven for classical reinforcement methods~\cite{zhang2011coordinated}, although not for deep models: as most works in the literature, we will use a benchmark to evaluate the performance of our scheme.} Formally, an \gls{ndpomdp} is identified by a 5-tuple, composed of a state space $\mathcal{S}$, an agent space $\mathcal{U}$, a joint action space $\mathcal{A}$, an observation space $\mathcal{O}$, and a reward map $r:\mathcal{S}\times \mathcal{A}\to\mathbb{R}^U$, where $U=|\mathcal{U}|$.

The complete system state $s$ is given by five matrices\edit{: a matrix for the current position of the \glspl{uav}, one matrix each for the map of the already discovered targets and obstacles, and one matrix each for the full map of targets and obstacles}. The positions of the \glspl{uav} are contained in the $2\times U$ matrix $\mathbf{X}$, while the features of the map are represented by the two $M \times M$ matrices $\bm{\Phi}$ and $\bm{\Omega}$, which contain the value $\phi(\mathbf{m})$ of each cell and the function $\omega(\mathbf{m})$ representing the location of the obstacles. \edit{Clearly, the maps with the full view of targets and obstacles are not initially known by the \glspl{uav}, which will then need to explore the area.} 

Furthermore, the \glspl{uav} do not know the features of cells that have not been explored: the observed features of the map are contained in the $F \times F$ observed value matrix $\hat{\mathbf{\Phi}}$, whose elements are equal to $\phi(\mathbf{m})$ if the cell has been explored and 1 otherwise, and the $F \times F$ observed obstacle matrix $\hat{\mathbf{\Omega}}$, whose elements are equal to $\omega(\mathbf{m})$ if the cell has been explored and 0 otherwise. The observation $o_u\in\mathcal{O}$ that is available to drone $u$ is then given by $\mathbf{X}_u$, $\hat{\bm{\Phi}}_u$, and $\hat{\bm{\Omega}}_u$, defined as the $F\times F$ subsets of $\mathbf{X}$, $\hat{\bm{\Phi}}$ and $\hat{\bm{\Omega}}$ centered in $\mathbf{x}_u$. 

\begin{figure}[t]
    \centering
    \begin{subfigure}{0.7\textwidth}
        \centering
        \includegraphics[width=\textwidth, trim=0cm 0cm 0cm 0.5cm,clip]{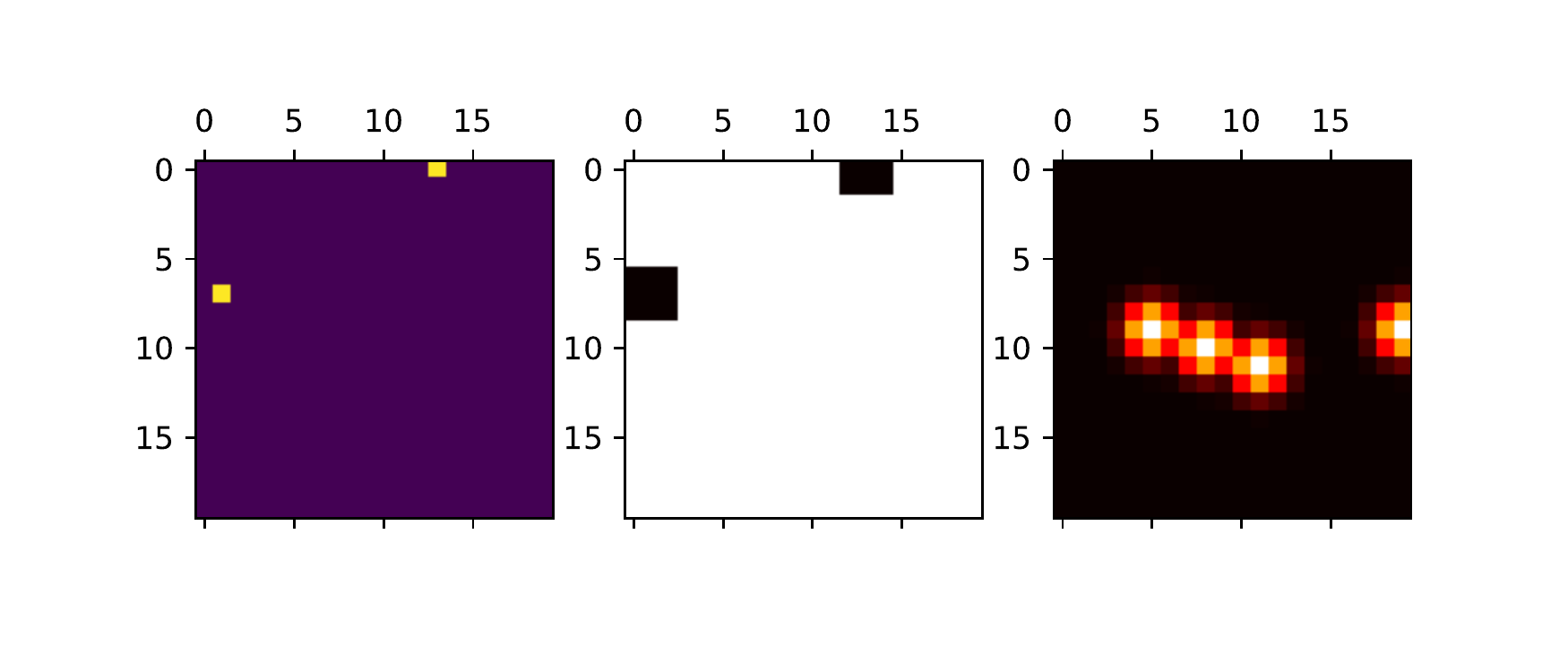}
        \label{fig:episode_start}
            \vspace{-2cm}
    \end{subfigure}
    \begin{subfigure}{0.7\textwidth}
        \includegraphics[width=\textwidth, trim=0cm 0cm 0cm 0.5cm,clip]{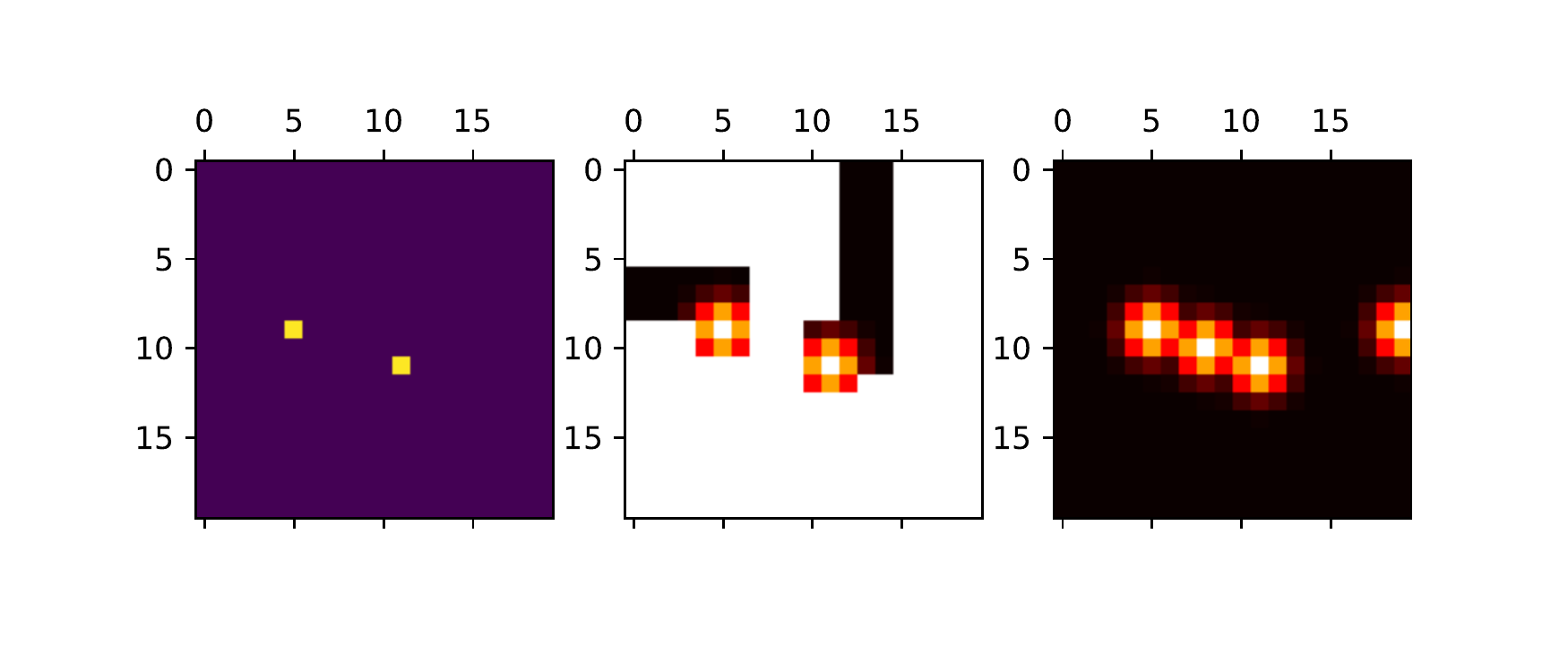}
        \label{fig:episode_end}
    \end{subfigure}
        \vspace{-1cm}
    \caption{Drone positions (left), known map (center), real map (right). Beginning (above) and end (below) of an episode.}
    \label{fig:episode_state}
\end{figure}

In our case, each \gls{uav} can either stay over the same cell or move to one of the four adjacent cells. However, obstacles are impassable in our environment definition, and the \glspl{uav} cannot move outside the map, so \glspl{uav} will simply stand in place if they attempt an action that violates the constraints. We define the action space $\mathcal{A} = \{(0,0)$, $(0,1)$, $(1,0)$, $(0,-1)$, $(-1,0)\}^U$. An action for the swarm is then a vector $\mathbf{a}\in\mathcal{A}$, which contains the individual \glspl{uav}' actions, denoted as $\mathbf{a}_u$ for drone $u$. We first define function $\edit{\nu}(\mathbf{x}_u,\mathbf{a}_u)$, which is 1 if the action is valid, i.e., it does not lead the \gls{uav} to fly outside the map or into an obstacle, \edit{and zero otherwise}:
\begin{equation}
    \edit{\nu}(\mathbf{x}_u,\mathbf{a}_u)=\begin{cases}
    1,& \text{if }\mathbf{x}_u+\mathbf{a}_u\in\mathcal{M}\wedge\omega(\mathbf{x}_u+\mathbf{a}_u)=0;\\
    0, & \text{otherwise}.
    \end{cases}
\end{equation}
The position of each drone is then updated in the following way:
\begin{equation}
    \mathbf{x}_u(t+1)=\mathbf{x}_u(t)+\mathbf{a}_u(t)\edit{\nu}(\mathbf{x}_u(t),\mathbf{a}_u(t)).
\end{equation}
Fig.~\ref{fig:episode_state} shows an example of the system state at the beginning and in an advanced stage of an episode, with two drones and four targets located in a $20 \times 20$ map with no obstacles (in this case, we set $F=M=20$). In particular, the drones' positions are shown on the left (in yellow), the observed value map is in the center, and the real value map is on the right. In the figure, darker cells are associated with lower values and brighter cells are associated with higher values. In the figure, if \edit{the communication range equals or exceeds the map side, i.e., } $F \geq M$, the observed state $o$ for all \glspl{uav} would correspond to the maps on the left and in the center. \edit{On the contrary,} if $F<M$, the observation for each \gls{uav} would \edit{include} a different portion of the map.
It is easy to see how the swarm gains knowledge during the episode, as the drones explore the map and look for targets.
In this case, the \glspl{uav} found two targets relatively quickly, and a significant portion of the grid remained unexplored.

We give reward 1 to a \gls{uav} if it is directly above a target, reward $-\theta$ if it tries to go outside the map or to position itself over an obstacle, reward $-\psi$ if it is in the same cell as another drone, and reward 0 in any other case. The \glspl{uav} \edit{will quickly} learn to avoid actions that would take them outside the map or make them crash into obstacles, so the exact value of $\theta$ does not affect the final performance, but the value of $\psi$ affects the distance that the drones try to keep from each other: if $\psi$ is low, the drones will get close to each other if the targets are very close. Naturally, if there is a collision risk when the drones are in the same cell, the value of $\psi$ should be high. The reward depends on $\mathbf{X}$, as well as on the action vector $\mathbf{a}$.

Indicating with $\mathbf{x}_u$ and $\mathbf{a}_u$ the position and action of drone $u$, we now define the collision variable $\chi_u(\mathbf{X}, \mathbf{A})$ as
\begin{equation}
    \chi_u(\mathbf{X}, \mathbf{A})=\max_{v\in(\mathcal{U}\setminus u)} \delta(\mathbf{x}_u+\mathbf{a}_u(t)\edit{\nu}(\mathbf{x}_u,\mathbf{a}_u) - \mathbf{x}_v \edit{- \mathbf{a}_v(t)}\edit{\nu}(\mathbf{x}_v,\mathbf{a}_v)).
\end{equation}
\edit{where $\delta(\mathbf{x})$ denotes a function that takes value $1$ if the vector $\mathbf{x}=0$, and zero otherwise.}
In short, $\chi_u(\mathbf{X}, \mathbf{A})$ has value 1 if one or more drones move to the same cell as drone $u$, and 0 otherwise.
The collision variable depends on the moves of other agents, so the problem is distributed.
The reward function for \gls{uav} $u$ in state $s$ if the swarm takes the joint action vector $\mathbf{A}$, denoted as $r_u(s,\mathbf{A})$, is given by:
\begin{equation}
\begin{aligned}
    r_u(s,\mathbf{A})= -\theta(1-\edit{\nu}(\mathbf{x}_u,\mathbf{a}_u))-\psi \chi_u(\mathbf{X},\mathbf{A})+(1-\chi_u(\mathbf{X},\mathbf{A}))\sum_{k=0}^{K-1}\delta(\mathbf{x}_u+\mathbf{a}_u-\mathbf{z}_k).
\end{aligned}
\end{equation}

In our model, the state transitions and the system observations are both deterministic; therefore, both the state evolution and the observation are not affected by random events, but only by the agents' decisions. We define a policy $\pi(\mathbf{a}_u|o_u)$ as the conditioned probability for user $u$ to take action $\mathbf{a}_u$ \edit{given} an observation $o_u\in\mathcal{O}$.
Under these assumptions, the goal of each drone $u$ is to find the policy $\pi^*$ that maximizes the cumulative expected future discounted reward $ R_{u}(\pi) = \mathbb{E}\left[\sum_{\tau=0}^{+\infty} \gamma^{\tau} r_{u,\tau}|o_u,\pi\right]$, where $\gamma\in[0,1)$ is a discount factor.

\subsection{Distributed Deep Q-Learning}


In this subsection, we will describe our \gls{ddql} approach to solve the problem defined above. For the sake of readability, \edit{in the following} we omit the $u$ subscript to indicate the agent whenever possible.
Each agent leverages a \gls{dqn}, i.e., a \gls{nn} that takes as input the last observation $o_t$ and returns the Q-values of the possible actions that can be taken, i.e., $Q(o_t, \mathbf{a}),\,\forall\mathbf{a} \in \mathcal{A}$.
In Q-learning, the function $Q(o, \mathbf{a})$ is an estimate of the expected long-term reward $R$ that will be achieved by choosing $\mathbf{a}$ as the next action and \edit{then} following the learned policy.
In our case, we maintain a single \gls{dqn} during the training phase, whose values are shared by all the agents. In this work, we follow the approach from~\cite{Mnih2015} and leverage a \textit{replay memory} to store the agent experience $e_t = $ $\left( o_t, a_t, r_t, o_{t+1}\right)$. Whenever the agent carries out a training step, a batch of $B_{\text{size}}$ elements is picked from the replay memory, allowing to separate the algorithm training from the experience acquisition. The replay memory is shared between the agents during a training phase, and a new batch is used to train the agent at every step. \edit{We highlight that, in our system, all agents are the same (single \gls{dqn}), and they need to generalize the problem from a limited number of states.
As it would be impossible for a single \gls{uav} to experience even just a non-negligible fraction of possible states in the training, shared replay is a critical factor in the network's generalization ability.
In particular, the experience replay is extremely valuable since it allows the system to improve the variety of the training samples by getting experience from the states seen by different agents.
In other scenarios, it may \edit{not be} convenient to exploit a shared memory, especially when the agents have to learn different tasks.}

\edit{Following the \gls{dqn} example from~\cite{Mnih2015}, we exploit the \textit{double Q-learning}
technique to remove biases from the Q-value estimation and speed up the algorithm's convergence~\cite{van2016deep}. This means that, during the training, we maintain a \emph{target network}, whose output $Q_t(o,a)$ is used to evaluate actions, and an \emph{update network}, whose output $Q_u(o,a)$ is used to select the policy. In particular, the bootstrap Q-value is computed as
\begin{equation}
    Q^{\text{new}}(o_t, a_t) = r_t + \gamma \max_a Q_t(o_{t+1}, a).
    \label{eq:double_q}
\end{equation}
The value $Q^{\text{new}}(o_t, a_t)$ is then used to perform backpropagation on the update network with a learning rate set automatically by the \gls{radam} optimizer~\cite{liu2019variance}, 
and every $n_q$ training steps the update network parameters are copied to the target network.}

\begin{figure}[t!]
\centering
\includegraphics[width=.7\textwidth]{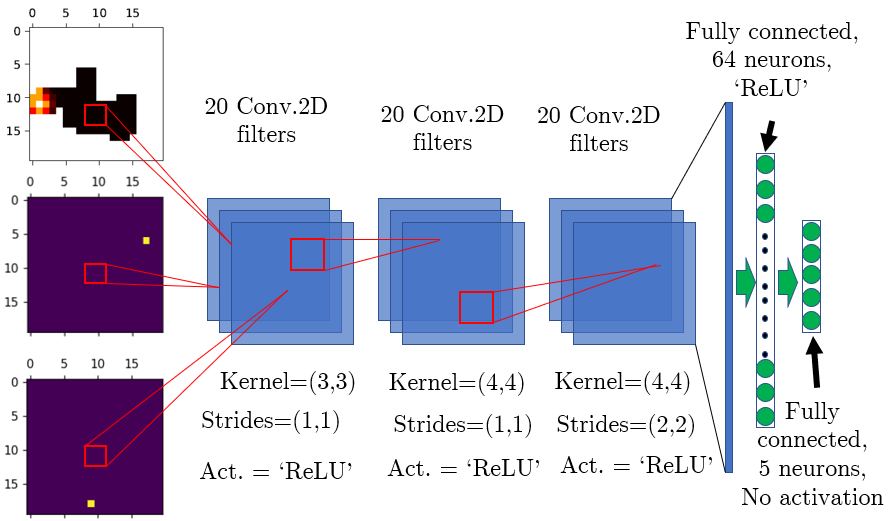}
\caption{Architecture of the \gls{dqn}.}
\label{fig:architecture}
\end{figure}


In our model, the observed state of the system for each agent can be represented by four $F \times F$ matrices, representing the agent position, the locations of the other agents, the value of explored cells, and the position of known obstacles. To simplify the state space, we consider matrices $\hat{\bm{\Phi}}$ and $\hat{\bm{\Omega}}$ jointly, by feeding the \gls{nn} with the matrix $\hat{\bm{\Phi}}-\rho\hat{\bm{\Omega}}$, \edit{where $\rho$ is a scalar parameter used to facilitate learning}.
Therefore, our system approximates the function $Q(o, a)$ by a \gls{cnn}, whose architecture is described in Fig.~\ref{fig:architecture}.
In particular, we consider a \gls{cnn} exploiting three convolutional layers followed by two fully-connected layers.
The dimension of the last layer is identical to the number of actions, so that each output element can be associated to a different action $a \in A$.

Hence, each agent provides training samples for the shared replay memory, which are then used in~\eqref{eq:double_q}, so that the \gls{cnn} output can converge to the Q-values $Q(o, a),\,\forall$ $a \in A$. We implement the well-known $\varepsilon$-greedy and softmax policies to allow the agents to explore the action space during the training phase, which is carried out by simulating a sequence of episodes.

\edit{\subsection{Computational complexity}}

\edit{We now discuss the computational complexity to perform one inference procedure with the neural network. \edit{We first analyze the complexity of fully-connected layers}. We \edit{denote by} $N_k$ the number of neurons in the general $k$-th layer. To go from layer $i$ to layer $i+1$, we need to compute the value of $N_{i+1}$ nodes, each of which takes $N_i$ multiplications followed by $N_i$ additions and one non linear function, thus involving $N_{i+1}(2N_i + 1)$ operations. }

\edit{We can then compute the complexity of one convolutional layer, as done in \cite{complexityCNN}, when neither batch normalization nor pooling layers are present. We denote with $(I_w, I_h, I_d)$ the shape of the input block. At layer $i$, we then have $K_i$ filters with kernels dimension $(W_i, H_i)$, stride $S_i$ (we use the same value along the two axes), and padding $P_i$. The shape of the resulting output block will be $\Huge( \frac{I_w + 2P_i - W_i}{S_i} + 1, \frac{I_h + 2P_i - H_i}{S_i} + 1,  K_i \Huge)$. The computation of each block's neuron here involves $W_i \times H_i \times I_d$ multiplications followed by the same number of additions (sum all elements plus the bias) and one non-linearity. The total number of calculations is then $\Huge( \frac{I_w + 2P_i - W_i}{S_i} + 1 \Huge) \times \Huge( \frac{I_h + 2P_i - H_i}{S_i} + 1 \Huge) \times  K_i \times \Huge( 2W_i H_i I_d + 1 \Huge)$.} 

\edit{If we consider the specific architecture of our \gls{nn} reported in Fig. \ref{fig:architecture}, the actual number of basic computations (multiplications, additions and non-linearities) are, respectively,  $440\,000$, $3\,704\,980$ and $628\,180$ for the three convolutional layers. The following fully-connected layers require $125\,504$ and $645$ computations, thus the total number of operations for one decision is $4\,899\,309$}.

\edit{This computational complexity allows \glspl{uav} to take decisions in time, as even embedded processors can deal with much more complex architectures in real time~\cite{bianco2018benchmark}. As the physical speed of the \glspl{uav} and the much more complex vision algorithms required to identify targets are the main limiting factors for the swarm, the \gls{ndpomdp} will be performed at a relatively slow pace, with timesteps in the order of several seconds.}

\section{Simulation settings}\label{sec:setup}

In this section, we describe the simulations by which we evaluated the performance of the designed system.
All the results are derived through a Monte Carlo approach, where multiple independent simulations are carried out to obtain reliable statistical data.
In particular, the algorithms' training is executed by carrying out a total of $N_e$ episodes for each studied scenario (sparse or cluster), where each episode is given by $N_s^t$ steps.
Training episodes are far longer than test episodes, which have length $N_s^p$, since the agents need to explore the map fully.

Before training, we initialize the replay memory by executing $N_e^m = 1000$ episodes of $N_s^t$ steps each, to allow agents to immediately start the learning procedure. If the episodes are too long, a lot of samples in which large portions of the map are already explored are added to the memory replay, and the agents will not learn properly how to move at the beginning of the episode, when the map is not explored. On the other hand, short episodes have the opposite problem, as the \glspl{uav} never learn to behave in the final parts of the episodes. A prioritized memory replay can solve this problem, but requires additional parameters. We then opted for adapting the episode length in the training phase. The even training episodes have 50 steps each, while the odd episodes have 150 steps. This alternating size prevents the replay memory \edit{from being} too skewed towards situations in which the map is almost completely explored or unexplored.

Moreover, we apply transfer learning to allow the agents trained in the sparse environment to quickly adapt to the cluster scenarios (or vice-versa); to this goal, additional $N_t$ training episodes are carried out. 
Finally, the performance of the proposed strategy is tested in a total of $N_p=500$ episodes for the \gls{ddql} system. The exploration rate $\varepsilon$ follows 2 different approaches, namely, $\epsilon$-greedy and softmax. In the former, a random action is chosen with probability $\epsilon$, while the best action, i.e., the action with the highest Q-value, is chosen with probability 1-$\epsilon$. The value of $\epsilon$ decreases to 0 at the end of the training, since no more exploration is needed. In the latter, at each time step the probability of each action $p_{i}$ is computed as the output of a softmax density function taking the Q-values as input. In this case, the temperature T decreases during the training, reducing the randomness during the selection of the actions:
\begin{equation}
    p_{i}=\frac{e^{\frac{q_i}{T}}}{\sum_{j=1}^{A} e^{\frac{q_j}{T}}}\ ,
\end{equation}
where $A=5$ is the number of actions that each drone can take. The training and testing processes are independently performed 5 times to verify the robustness of the \gls{ddql} scheme. The complete simulation settings are reported in Tab \ref{tab:sim_param}. 

\begin{table}[t]\centering
    \footnotesize
	\begin{tabular}[c]{ccc}
		\toprule
		Parameter & Value & Description \\
		\midrule
		$M$ &\{20, 24, 30, 40, 50\} & Map size \\
		$F$ & 20 & Observed map size \\
		$U$ & $\{2, 3\}$ & Number of \glspl{uav} \\
		$K$ & 4 & Number of targets \\
		$\sigma^{2}$ & 1 & Targets variance \\
		$\zeta$ &  3 & Field of View \\
		$\eta$ & \{0,0.1\} & Obstacle frequency\\
		$d_{\text{sparse}}$ & 8 & Minimum target distance (sparse scenario)\\
		$\theta$ & 1 & Obstacle/outside penalty \\
		$\psi$ & 0.8 & Collision penalty \\
		$\rho$ & 0.2 & Obstacle value\\
		$\gamma$ & 0.9 & Discount factor \\
		$\alpha$ & Chosen by \gls{radam} & Learning rate \\
		$N_e$ & $\{250, 750, 1000, 3000\}$ & Training episodes \\
		$N_s^t$ & $\{50,150\}$ & Steps per training episode \\
		$N_{s}^p$ & 40 & Steps per test episode \\
		$N_t$ & $\{125, 250, 375,750\}$ & Transfer learning episodes \\
		$N_p$ & 100 (LA), 500 (DDQL) & Test episodes \\
		$P_{\text{tx}}$ & 20 dBm & Communication power\\
		$N_0$ & -76 dBm & Noise floor\\
		$h$ & 40 m & \gls{uav} height\\
		$R_c$ & $2/3$ & Coding rate\\
		\bottomrule
	\end{tabular}
	\vspace{0.1cm}
	\caption{Simulation settings.}
	\label{tab:sim_param}\vspace{-0.8cm}
\end{table}

To assess the performance of our \gls{ddql} scheme, we compare it with a heuristic strategy inspired by \gls{mpc}, by which drones can explore the map and reach the targets.
Such a strategy is named \emph{look-ahead} and is used as a benchmark for our analysis. The look-ahead strategy tries all possible combinations of future actions and looks at the possible future rewards, as its name suggests. In order to define it, we first define the look-ahead reward $r_u^{(\ell)}(\mathbf{X},\mathbf{a})$ as:
\begin{equation}
    r_u^{(\ell)}(\mathbf{X},\mathbf{a}) = \begin{cases} \frac{\hat{\phi}(\mathbf{x}_u+\mathbf{a}_u)}{\xi(\mathbf{x}_u+\mathbf{a}_u)} &\text{if } \edit{\nu}(\mathbf{x}_u, \mathbf{a}_u)=1;\\
    -\infty &\text{otherwise,}
    \end{cases}
\end{equation}
where $\xi(\mathbf{x})$ is the number of \glspl{uav} located in $\mathbf{x}$. The look-ahead strategy never goes outside the map or on obstacles. To decide its next action, each drone $u$ tries to maximize its expected cumulative reward over the following $n_{\ell}$ steps, assuming that none of the other drones move. 
Practically, the look-ahead strategy makes each drone select the action $\mathbf{a}^*$ that maximizes 
\begin{equation}
\label{eq:la_action}
    \max_{\mathbb{A} \in  \mathbf{\tilde{\mathcal{A}}}^{n_\ell} }
    \sum_{i=0}^{n_\ell-1}
    r_u^{(\ell)}\left(\mathbf{X}+\sum_{j=0}^{i-1}\mathbf{A}^j,\mathbf{a}^i\right),
\end{equation}
where
$\tilde{\mathcal{A}}^{n_\ell}$ is the set of ordered sequences $\mathbb{A}$ of action vectors $\mathbb{A}^0$, $\mathbb{A}^1$, ..., $\mathbb{A}^{n_\ell-1}$,
so that $\hat{\mathbf{a}}^0_u = a^*$ and $\mathbf{a}^i_v = (0,0),\,\forall$ $i \in \{0,...,n-1\}, v \neq u$, i.e., the set of possible move sequences of $u$ while the other \glspl{uav} are static. If several action sequences have the same expected reward, the look-ahead strategy will choose one of them randomly.
At the beginning of an episode, each drone $u$ assumes that all the map values $\phi(\mathbf{m})$ outside its \gls{fov} are equal to $1$; therefore, look-ahead forces $u$ to continuously explore the map. However, as soon as it finds a target, $u$ will hover over the target center. The target is then eliminated from the other agents' value maps, as it is already covered by a \gls{uav}.
We highlight that the performance of look-ahead mainly depends on the $n_{\ell}$ parameter: as it increases, drones can make more foresighted decisions, but at a greater computational cost. In addition, the number of targets in the map also plays a key role in determining the computational performance: when more targets are present, we have to check whether other agents are on a target more often, in order to remove it from the map of available targets. As the look-ahead strategy is computationally expensive, $N_p$ for it was set to 100.

\edit{Finally, we also consider a scenario with a realistic communication model, in which the broadcast messages sent by each \gls{uav} at every step might be lost due to the wireless channel impairments.
We used the path loss and shadowing model from~\cite{liu2019measurement}, based on actual measurements from air-to-air communications, and considering a Rayleigh fading model with an error correction code with rate 2/3.
As the simulation results will show, the physical size of the cells in the map is a critical parameter when \glspl{uav} communicate directly with each other (and not through the network infrastructure on the ground).
In particular, increasing the size of the cells will impair \edit{the} performance because of communication range issues: the model has an error probability of 50\% at approximately 110 m, corresponding to 11 cells if a cell side is 10 m and 5 cells if the side is 20 m.}

\section{Simulation results}\label{sec:results}

\edit{In what follows, we evaluate the performance of our approach in various scenarios with different characteristics.} 

\subsection{Training analysis}

\begin{figure}[t!]
\centering
\includegraphics[width=.7\textwidth,trim=0cm 0.5cm 1.5cm 2cm,clip]{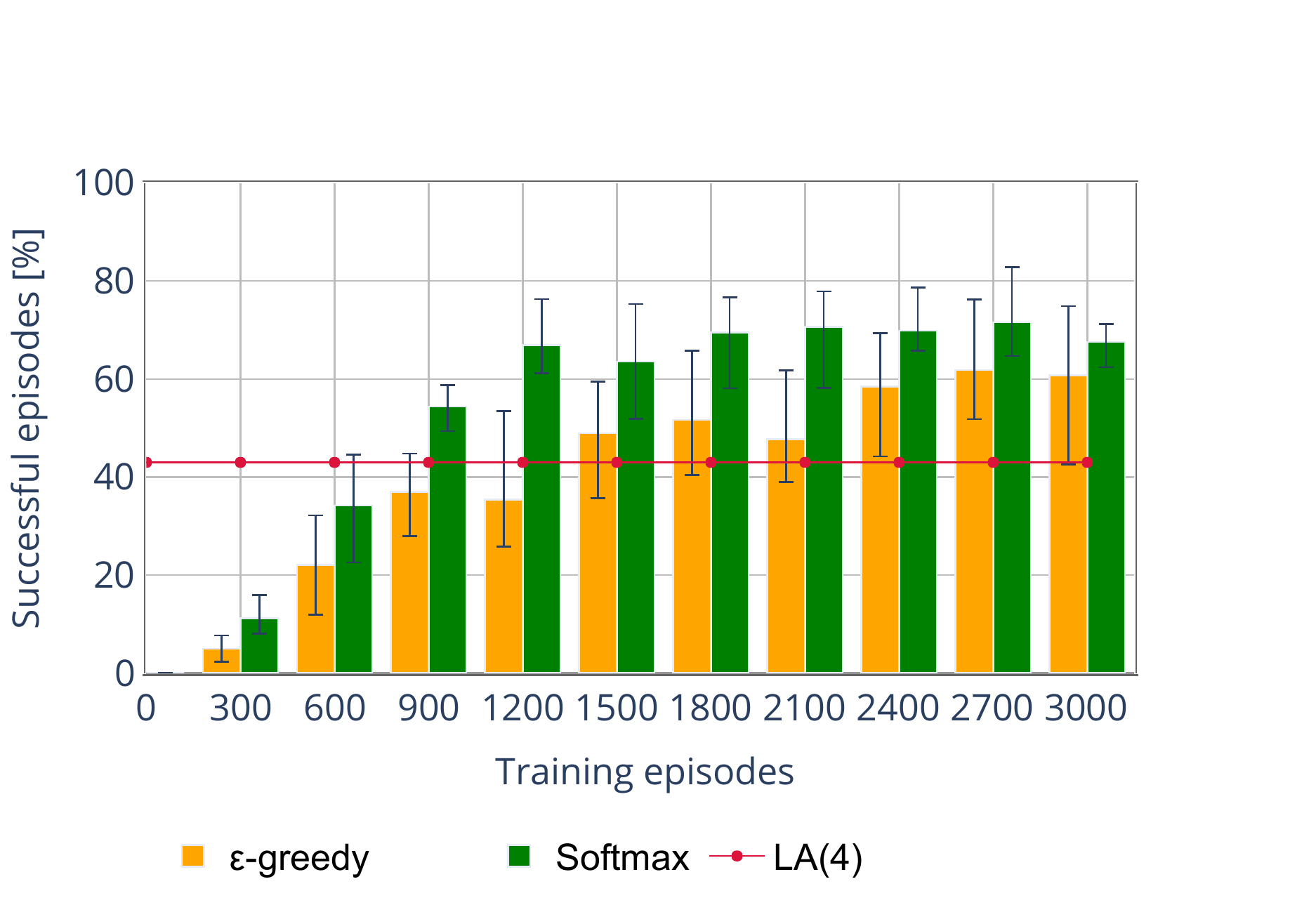}
\caption{Success probability over the training phase in the \emph{cluster} scenario with 2 UAVs.}
\label{fig:training_cluster}
\end{figure}

\edit{We first consider a scenario with 2 \glspl{uav} and 4 targets in a $20\times20$ map.
In particular, we perform multiple training phases of different duration; the longer training includes 3000 episodes, for a total of 300,000 training samples, which ensures that all our algorithms achieve convergence.}
The look-ahead approach is abbreviated as LA(4), as we set $n_\ell=4$.
This already had a significant computational cost, and in our simulation each look-ahead decision takes approximately 15 times longer than running a trained \gls{ddql} agent.
\edit{We do not consider $n_\ell>4$, since the computational cost of such a technique becomes excessive with limited performance gains: without coordination among the \glspl{uav}, which requires a prediction of the movements of other drones in the swarm, there is a limit on the performance of the swarm even with an infinite horizon. In some brief tests (which had to be on maps of a limited size due to the computational complexity of LA with a longer horizon), we noticed that LA(8) and even LA(12) show limited gains over LA(4), as the biggest factor in determining the speed at which the \glspl{uav} find the target becomes the coordination of the swarm once the horizon reaches 3 or 4 steps.}

\begin{figure}[t!]
\centering
\includegraphics[width=.7\textwidth,trim=0cm 0.5cm 1.5cm 2cm,clip]{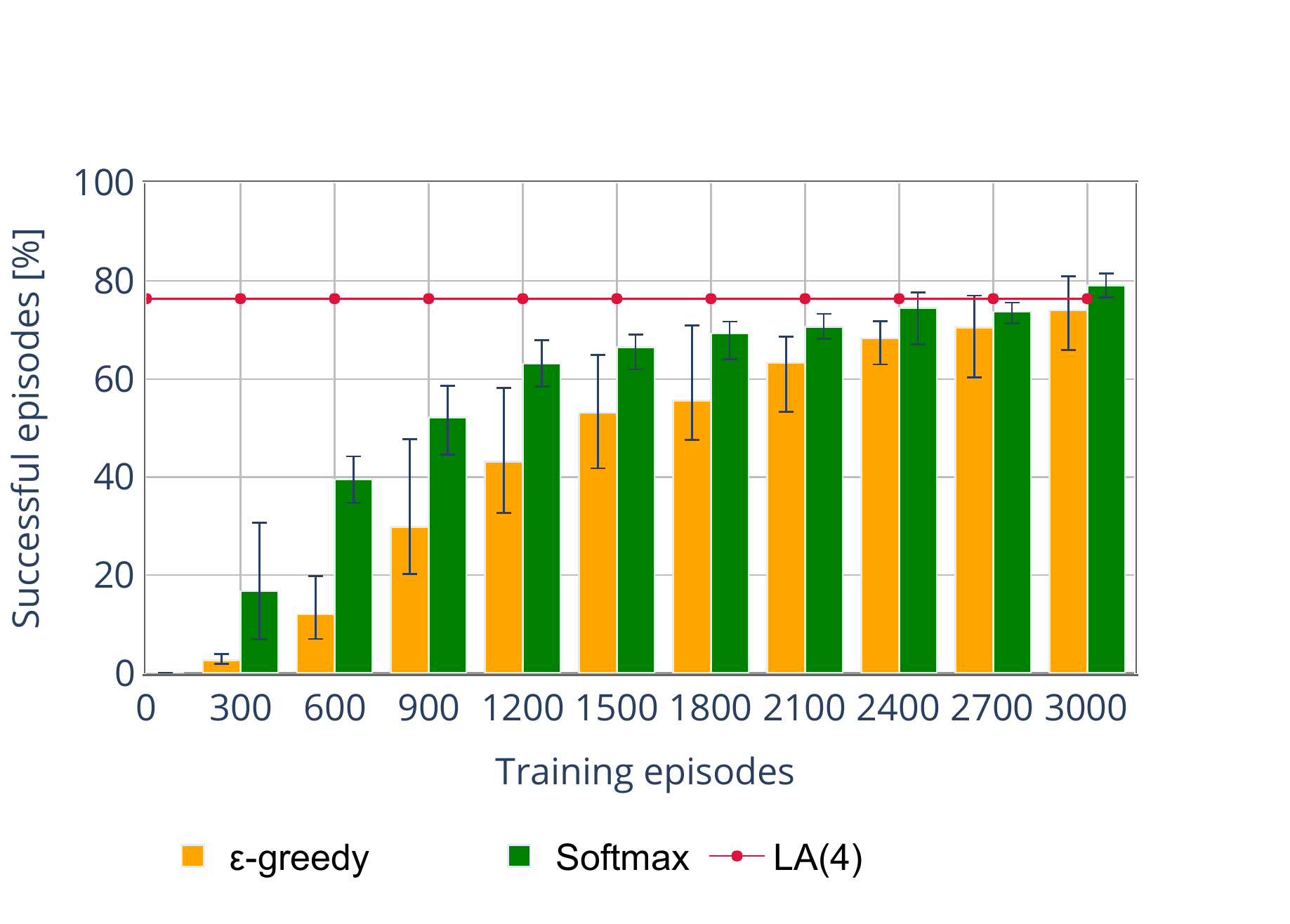}
\caption{Success probability over the training phase in the \emph{sparse} scenario with 2 UAVs.}
\label{fig:training_sparse}
\end{figure}

Fig.~\ref{fig:training_cluster} shows the success probability in the \emph{cluster} scenario as a function of the training set size and of the considered exploration profile and approach. \gls{ddql} combined with the softmax approach catches up with LA(4) in less than 900 training episodes, converging to a success probability between 0.65 and 0.7. The $\epsilon$-greedy approach has a lower final performance and requires more time to converge with respect to the softmax profile.
The error bars show the best and worst results over 5 test phases, showing that the performance improves as the \glspl{uav} gain more experience.
The performance boost over the look-ahead approach is due to the \gls{ddql} scheme's ability to exploit the correlation among the target positions, quickly finding the other targets after the first one has been spotted. 
Instead, in the \emph{sparse} scenario, the final performance of \gls{ddql} is similar \edit{to that} of LA(4), as Fig.~\ref{fig:training_sparse} shows.
In general, both \gls{ddql} and LA(4) have more success than in the cluster scenario, as finding the scattered targets is easier than finding clusters in the limited duration of an episode.

\begin{figure}[t]
    \centering
    \begin{subfigure}{0.45\textwidth}
        \centering
        \includegraphics[width=\textwidth, trim=0cm 4cm 1cm 0cm,clip]{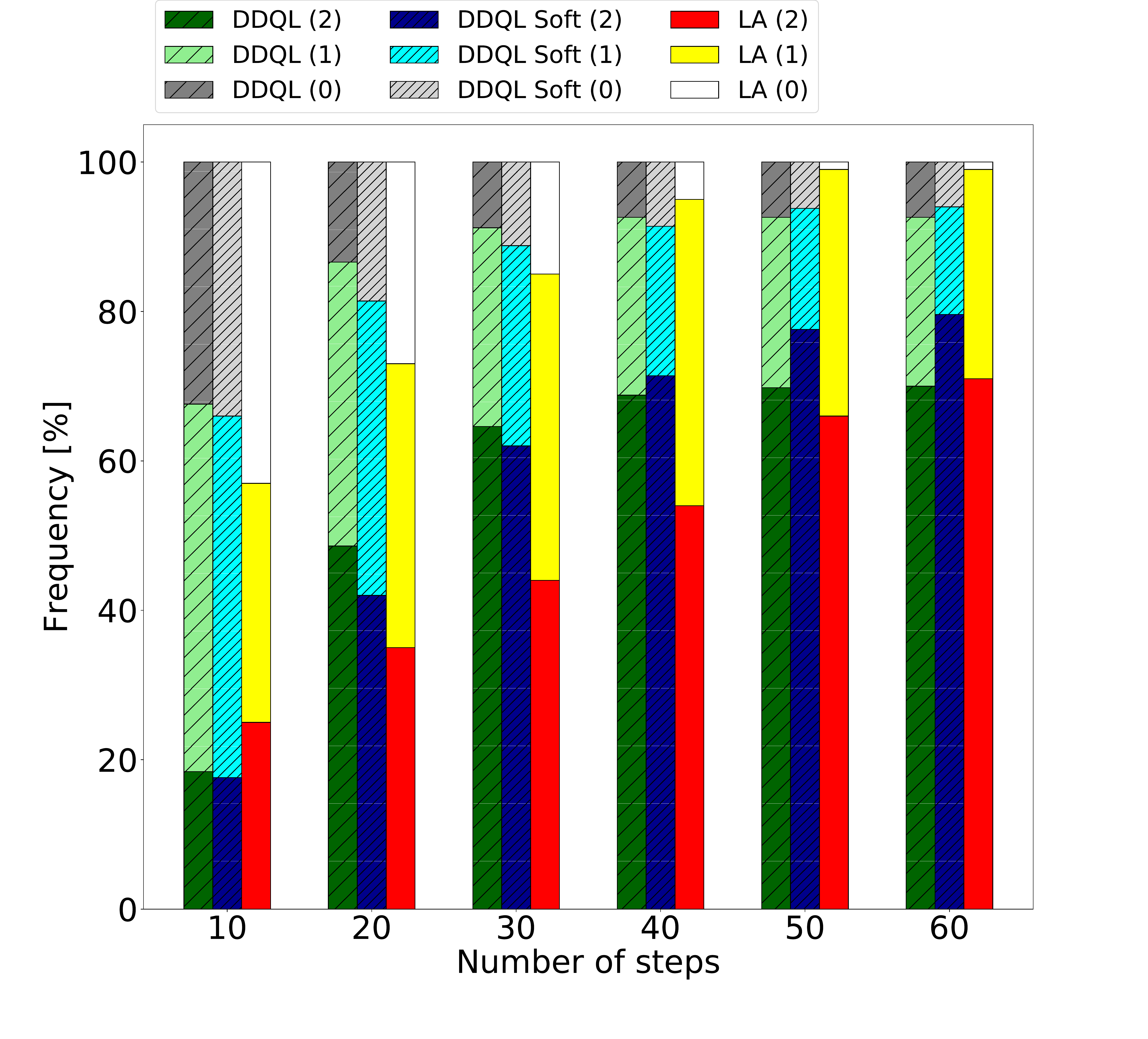}
        \caption{Cluster scenario}
                \label{fig:cluster_state}
    \end{subfigure}
    \begin{subfigure}{0.45\textwidth}
        \includegraphics[width=\textwidth, trim=0cm 4cm 1cm 0cm,clip]{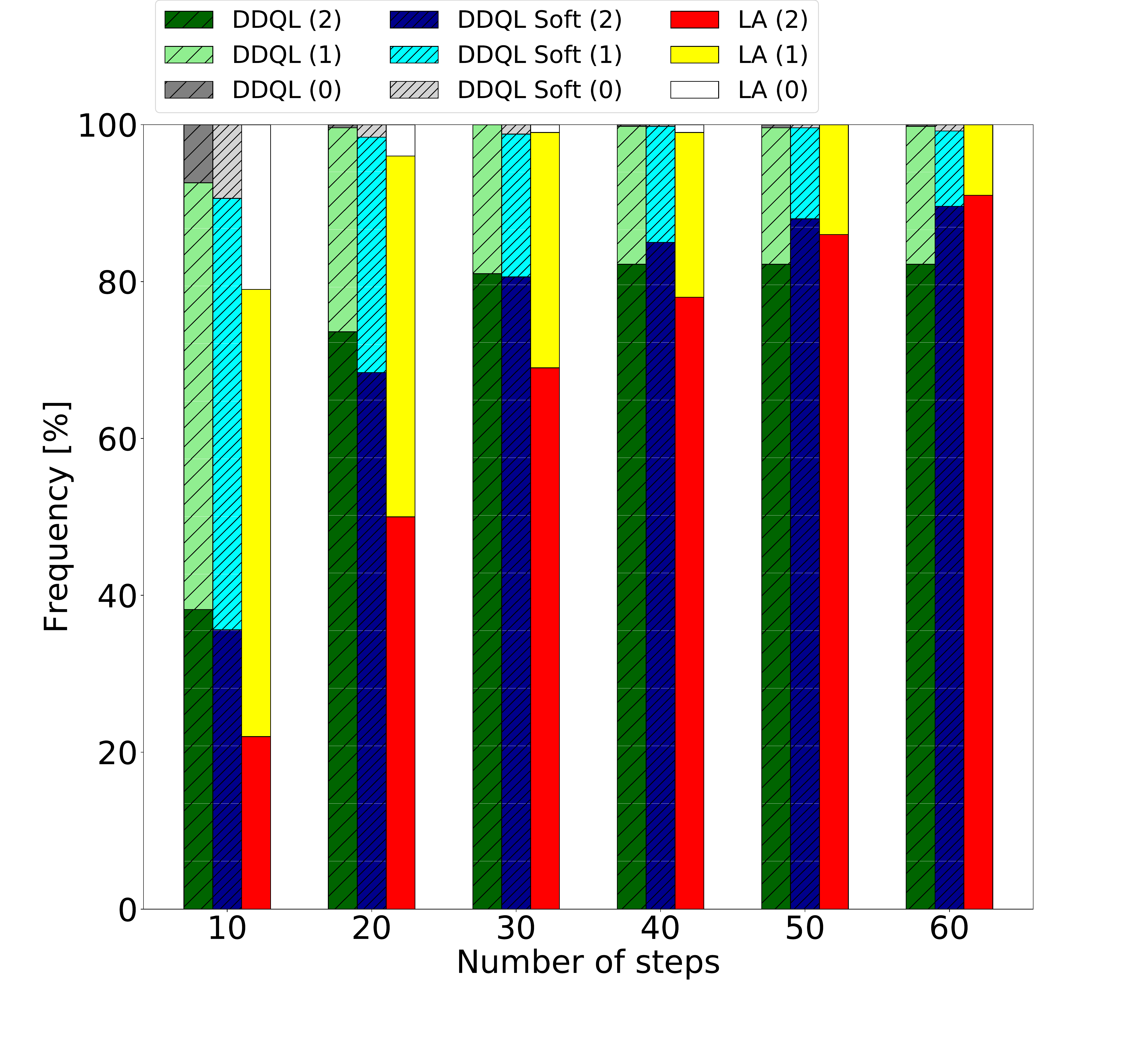}
        \caption{Sparse scenario}
                        \label{fig:sparse_state}
    \end{subfigure}
    \caption{\edit{The bars indicate the probability mass distribution of the number of \glspl{uav} that successfully accomplish their task (i.e., hover upon a target) by the end of the episode, when varying the duration of the episode. Each group of bars refers to the performance achieved by \gls{ddql} (with and without softmax) and by LA, in the Cluster (a) and Sparse (b) scenarios, with a total of 4 targets and 2 \glspl{uav}}}.
    \label{fig:prob_bars}
\end{figure}

\subsection{Success rate vs time}

\edit{The next set of results refer to the performance of the strategy learned by the proposed framework.  Fig.~\ref{fig:prob_bars}} reports the probability of one or both drones reaching the target as a function of the number of steps. \edit{Therefore, the figure shows the trade-off  between the time needed by \glspl{uav} to accomplishing their task and the success rate. } In the cluster scenario (Fig.~\ref{fig:cluster_state}), \gls{ddql} is much faster than LA, but its performance peaks out, and after 40 steps the probability of the \glspl{uav} reaching their targets does not change significantly. \edit{Indeed, we observed that, in certain cases,} when a drone reaches the target, but the other one is far from any feature of the map, the latter can end up staying in place, as its Q-values for that scenario are not precise and all actions have a similar (low) value. This almost never happens before the first \gls{uav} reaches its target, \edit{since the change in the system state due to the movement of one \gls{uav} is generally enough to make the other \gls{uav} move.} This is not a problem for LA, whose performance keeps rising \edit{with time}; in the sparse scenario (Fig.~\ref{fig:sparse_state}), LA even ends up reaching more targets than \gls{ddql} after 50 steps. The solution we found to avoid \edit{this roadblock} is simply to maintain a low softmax temperature $\tau=0.1$ even during the test phase: the bar chart shows that the \gls{ddql} Soft system is slightly slower than the greedy \gls{ddql} at the beginning, but it can avoid getting stuck. This randomization allows the agent to get out of loops, as sometimes a random sub-optimal action will change the state and allow it to reconsider, while the greedy system will keep performing the same action and remain in the same state. LA essentially does the same, randomizing its action when it is unsure which one is the best.

\begin{figure}[t!]
\centering
\includegraphics[width=.7\textwidth,trim=1cm 1cm 1cm 2cm,clip]{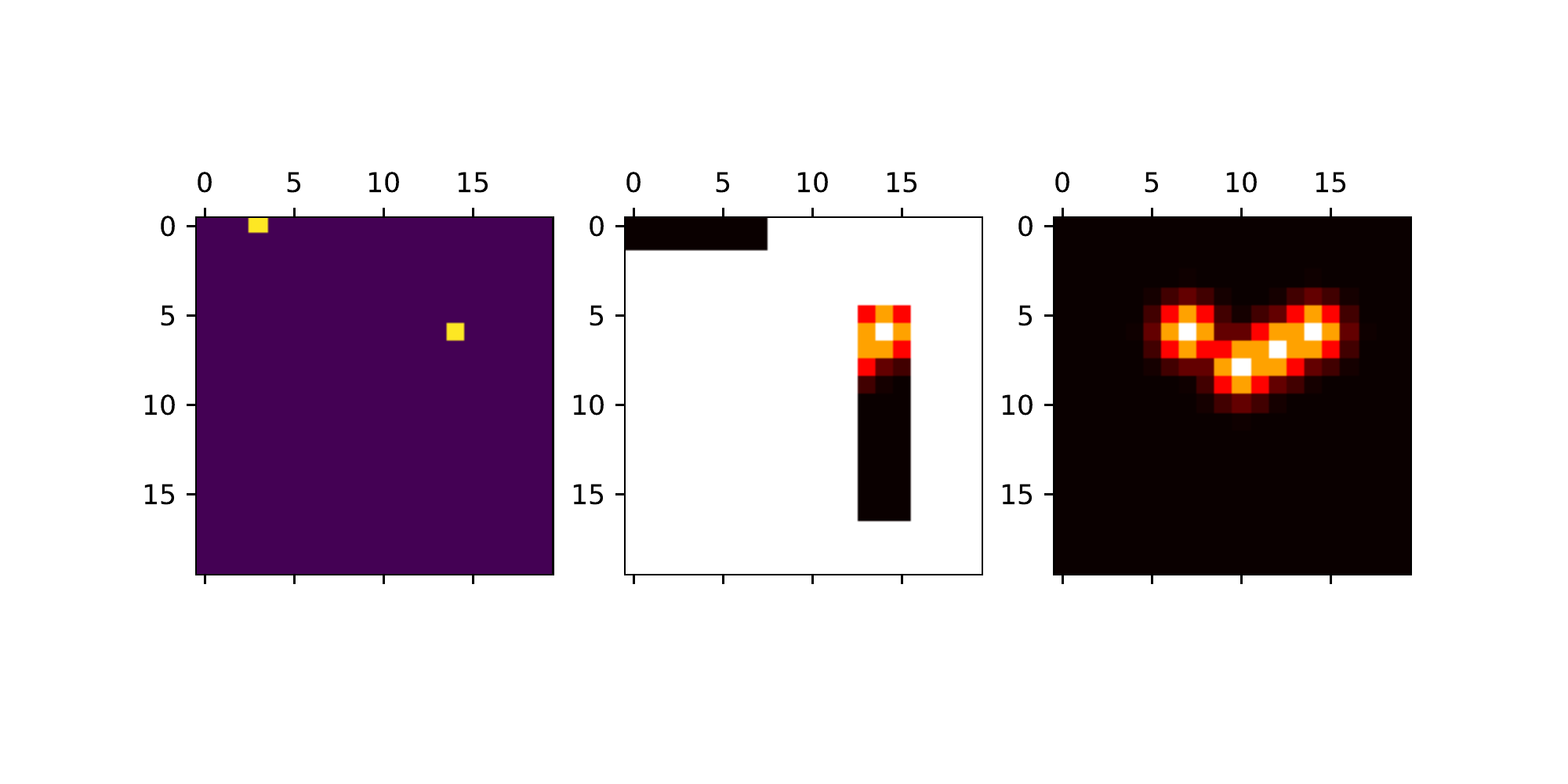}
\caption{Example of an episode where the second \gls{uav} is not able to reach the cluster}
\label{fig:Bad_situation}
\end{figure}
Fig. ~\ref{fig:Bad_situation} shows one such situation: as one \gls{uav} has reached its target, while the other is far from any identified target, its Q-values will be very similar to each other, and some of the time it will stay motionless or move in small loops, as its state never changes. The fact that most of the map is still unexplored increases the probability of the \gls{uav} getting stuck, as it will have limited information and its Q-values will be very similar. In the following, all the results are referred to the \gls{ddql} Soft system with $\tau=0.1$ unless otherwise stated.

\begin{figure}[t!]
\centering
\includegraphics[width=.7\textwidth]{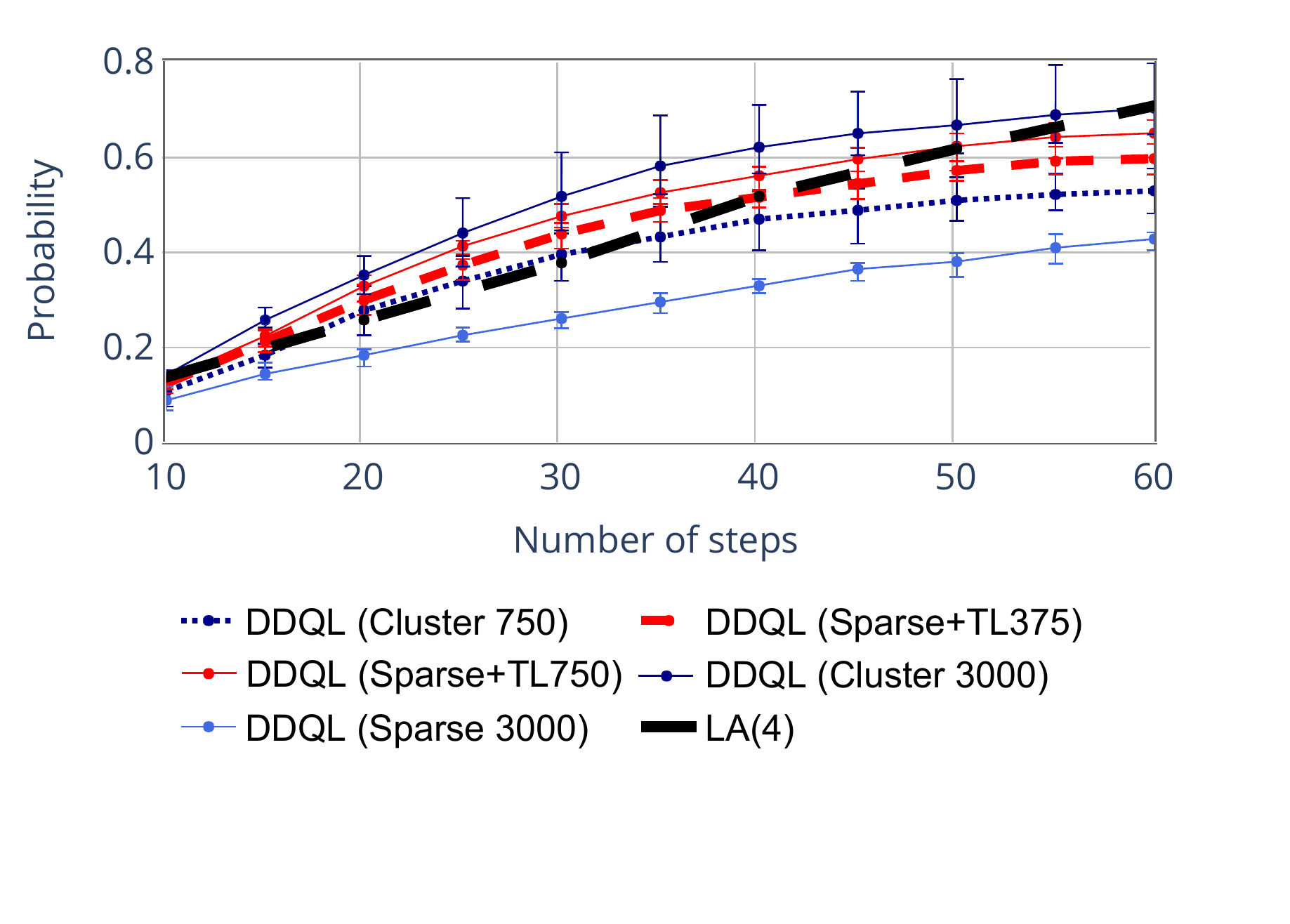}
\caption{CDF of the episode duration for different algorithms in the cluster scenario with 2 \glspl{uav}.}
\label{fig:transfer_cluster_2d}
\end{figure}

\begin{figure}[t!]
\centering
\includegraphics[width=.7\textwidth,trim=0cm 0.5cm 1cm 2cm,clip]{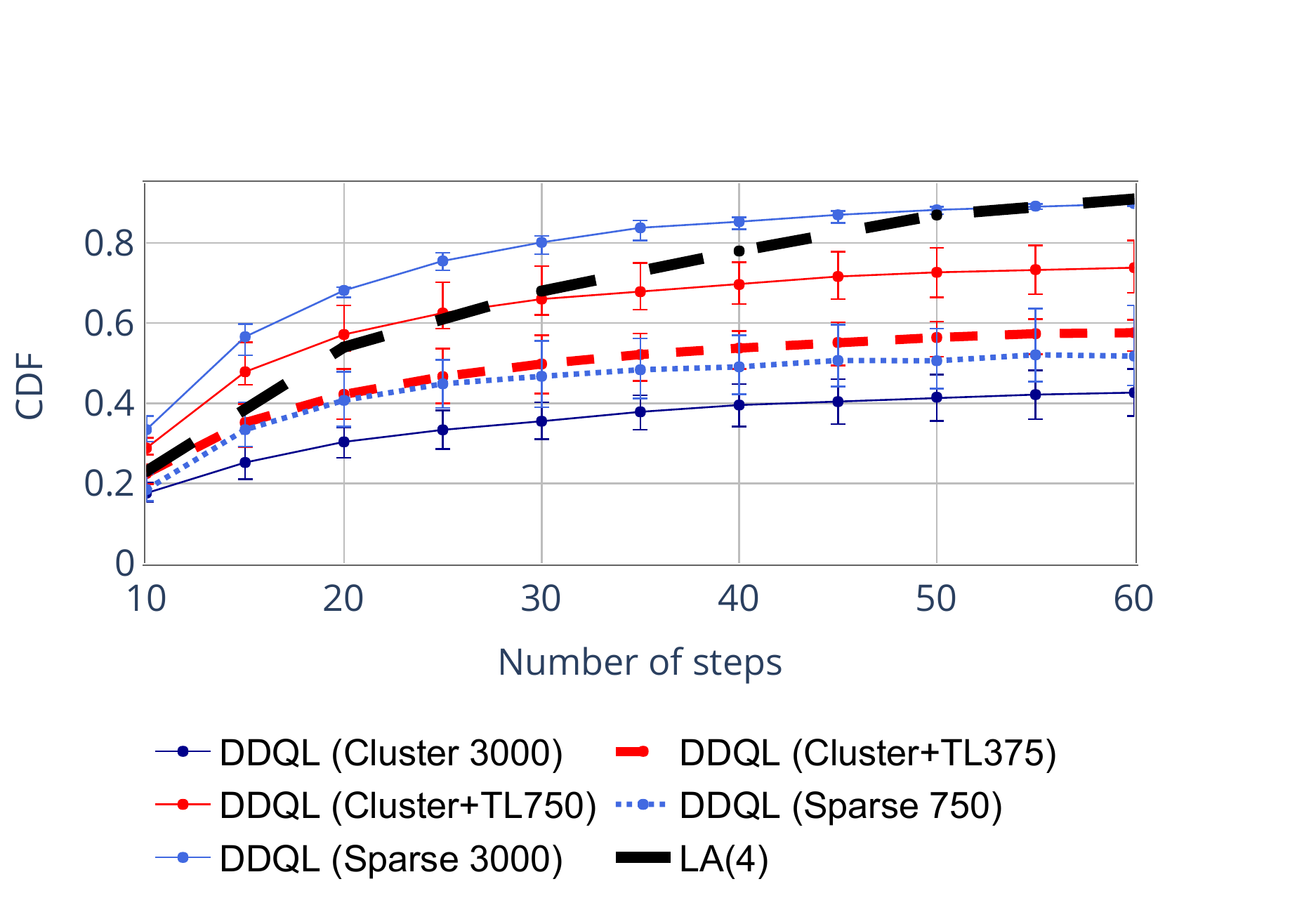}
\caption{CDF of the episode duration for different algorithms in the sparse scenario with 2 \glspl{uav}.}
\label{fig:transfer_sparse_2d}
\end{figure}



\subsection{Adaptability and transfer learning}

\edit{Here we investigate the adaptability of the proposed \gls{ddql} scheme, and the potential of the transfer learning paradigm.
\edit{The} latter involves the execution of an additional training phase in a different scenario than the one seen during the initial training.
To this end, we consider a common  target  scenario, i.e., cluster (or sparse),  and  compare  the  results  achieved  when  using strategies learned in the other domain, i.e., sparse (or cluster).
More specifically, we consider the following cases: 
\begin{itemize}
\item "Cluster $N_e$": training on $N_e$ episodes in the cluster scenario;
\item  "Sparse $N_e$ ": training on $N_e$ episodes in the sparse scenario; 
\item "Cluster+TL $N_t$": pre-training on $N_e$ = 3000 episodes in the cluster scenario, followed by an additional training of $N_t$ episodes in the target scenario. 
\item "Sparse+TL $N_t$": pre-training on $N_e$ = 3000 episodes in the sparse scenario, followed by an additional training of $N_t$ episodes in the target scenario. 
\end{itemize}}

\edit{In Fig. \ref{fig:transfer_cluster_2d}} we show the \gls{cdf} of the \edit{\textit{episode duration}, defined as} the time until \edit{all the} drones reach targets or the testing episode limit \edit{(here fixed to 60 steps)} is reached. \edit{We also report the results for LA with four steps, LA(4), as a benchmark.} \edit{Each point is hence the probability that all drones have accomplished their task by a  given number of steps.}

\edit{We observe that, as expected, the Cluster strategy achieves the highest success probability with a limited number of steps. LA(4) can equal its performance only when the episode duration reaches the limit of 60 steps (i.e., in less than 30\% of the cases).} Instead,
750 episodes of training in the cluster scenario are \edit{not sufficient to outperform LA(4), but actually enough to} outperform a model trained in the sparse scenario.  
However, \edit{a short} retraining \edit{of such model} in the correct (cluster) scenario allows the algorithm to get a significant performance boost, \edit{outperforming LA(4) and getting very close to the performance of the Cluster 3000  model, which is fully trained in the correct scenario and with more than twice the number of episodes. }

\begin{figure}[t!]
\centering
\includegraphics[width=.7\textwidth,trim=0cm 0.5cm 1cm 2cm,clip]{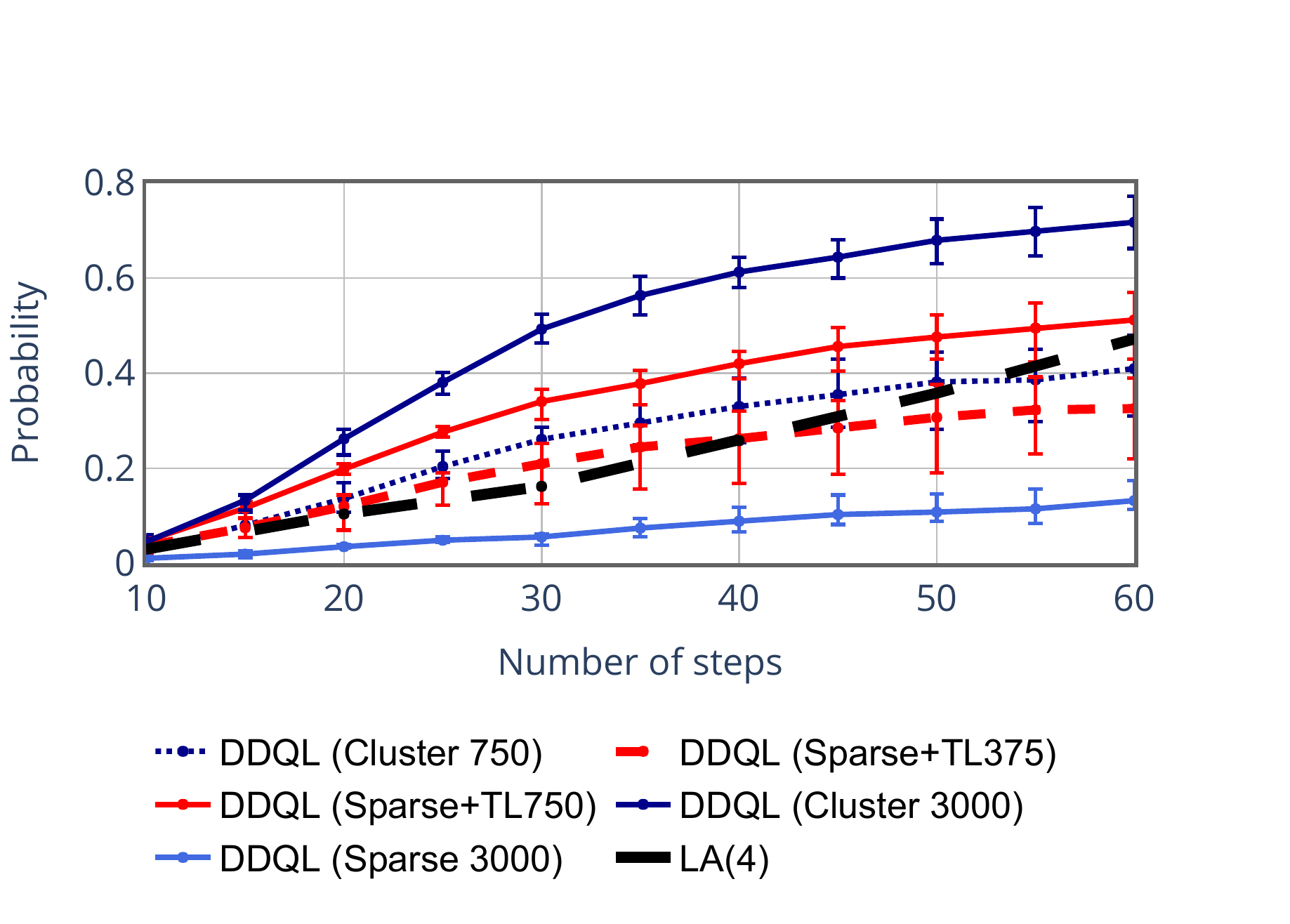}
\caption{CDF of the episode duration for different algorithms in the cluster scenario with 3 \glspl{uav}.}
\label{fig:transfer_cluster_3d}
\end{figure}

\begin{figure}[t!]
\centering
\includegraphics[width=.7\textwidth,trim=0cm 0.5cm 1cm 2cm,clip]{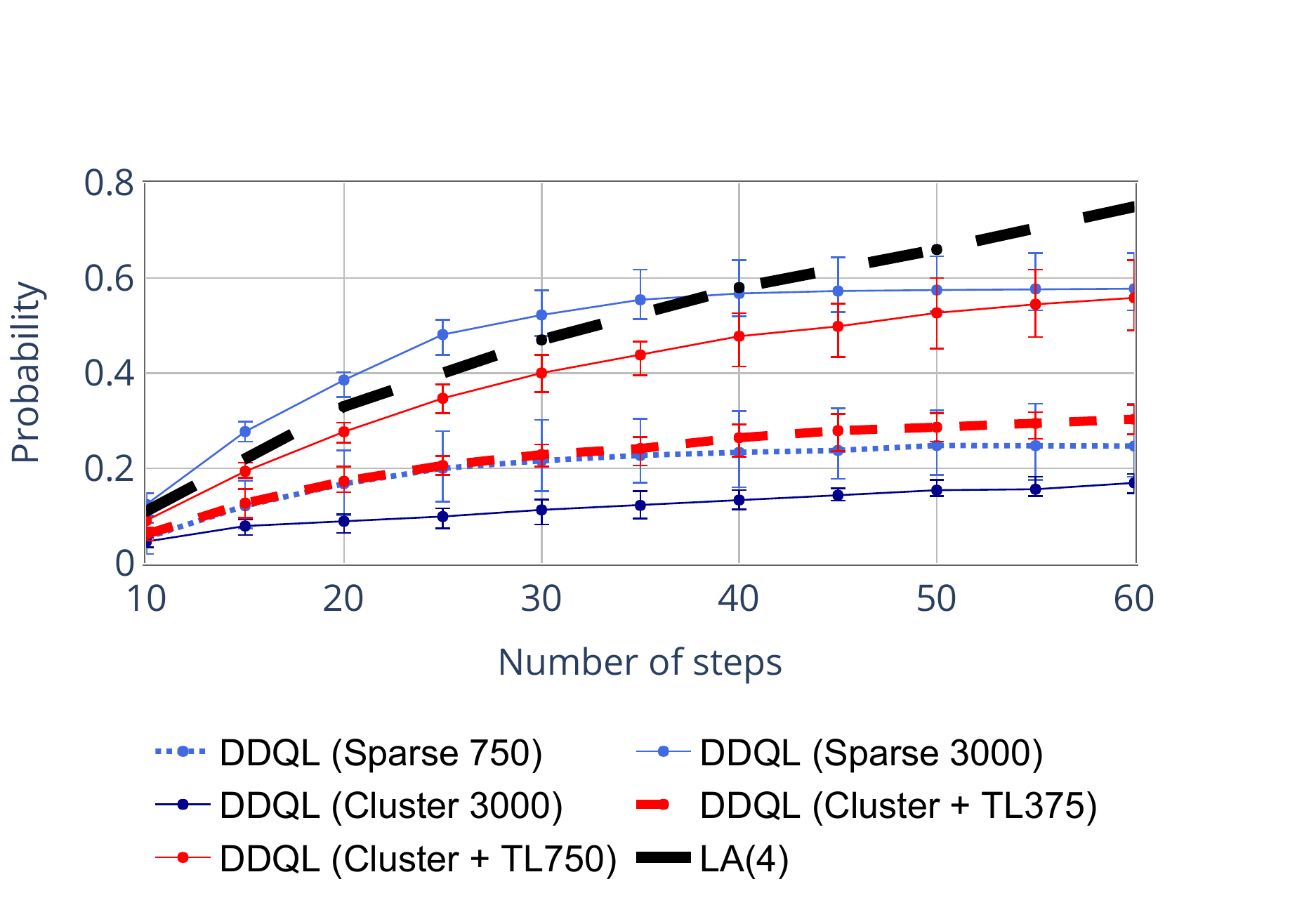}
\caption{CDF of the episode duration for different algorithms in the sparse scenario with 3 \glspl{uav}.}
\label{fig:transfer_sparse_3d}
\end{figure}

We repeated the experiment \edit{by swapping the role of the sparse and cluster scenarios, and changing the number of episodes during the training phase, as reflected in the legend of Fig. \ref{fig:transfer_sparse_2d}, which reports the results.  Also in} this case, \edit{LA(4)} meets the performance of \gls{ddql} \edit{only for episodes of} 60 steps, \edit{i.e., in less than 15\% of the cases.} Transfer learning is \edit{again} very effective, as a 750 episode re-training significantly boosts the baseline performance compared to starting from scratch. \edit{We highlight that, in general, the number of steps necessary to reach the targets is comparatively lower than in the previous scenario since, as already discussed, it is easier for \glspl{uav} to find targets in} the sparse scenario.

\edit{In Fig. \ref{fig:transfer_cluster_3d} and Fig. \ref{fig:transfer_sparse_3d}} we show the results for a scenario with 3 \glspl{uav}: in both cases transfer learning is effective, but the performance is lower in the sparse scenario than in the cluster one. In this case, the risk of getting stuck is increased and the algorithm needs more training to perform effectively in all maps.

\subsection{Obstacles}

\begin{figure}[t]
    \centering
    \begin{subfigure}{0.7\textwidth}
        \centering
        \includegraphics[width=\textwidth, trim=0cm 0cm 0cm 0.5cm,clip]{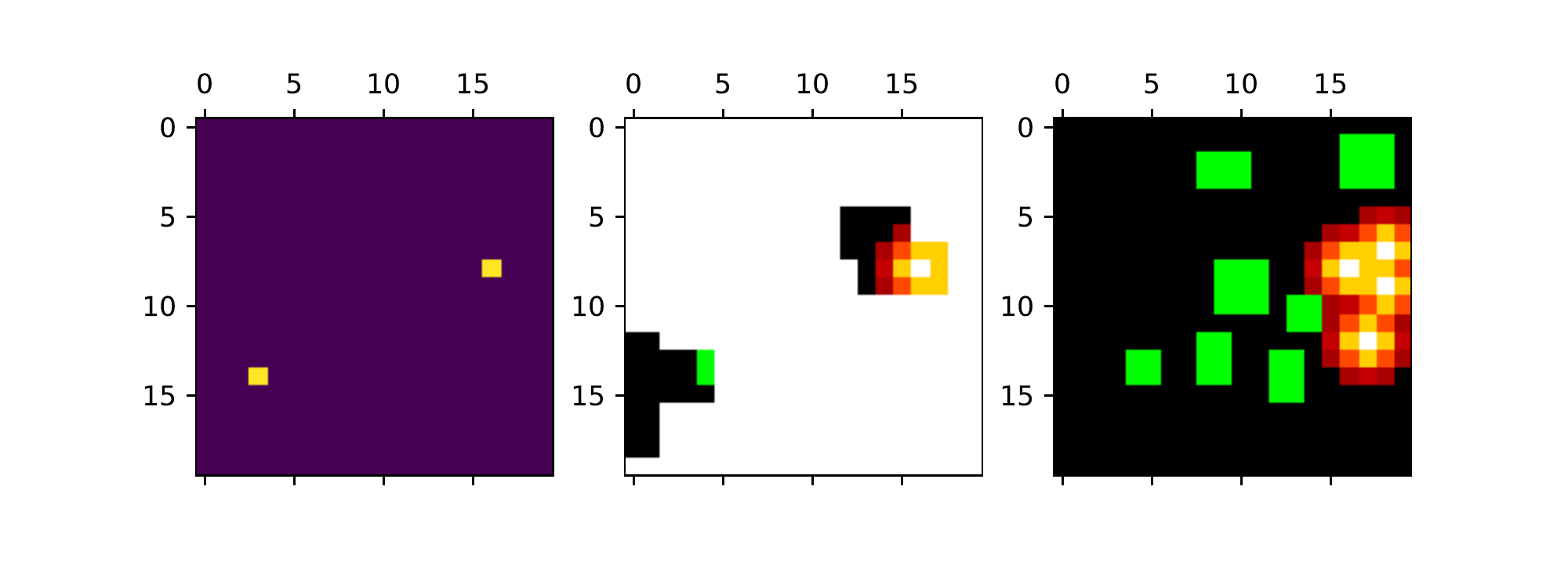}
        \label{fig:episode_start_obs}
            \vspace{-1cm}
    \end{subfigure}
    \begin{subfigure}{0.7\textwidth}
        \includegraphics[width=\textwidth, trim=0cm 0cm 0cm 0.5cm,clip]{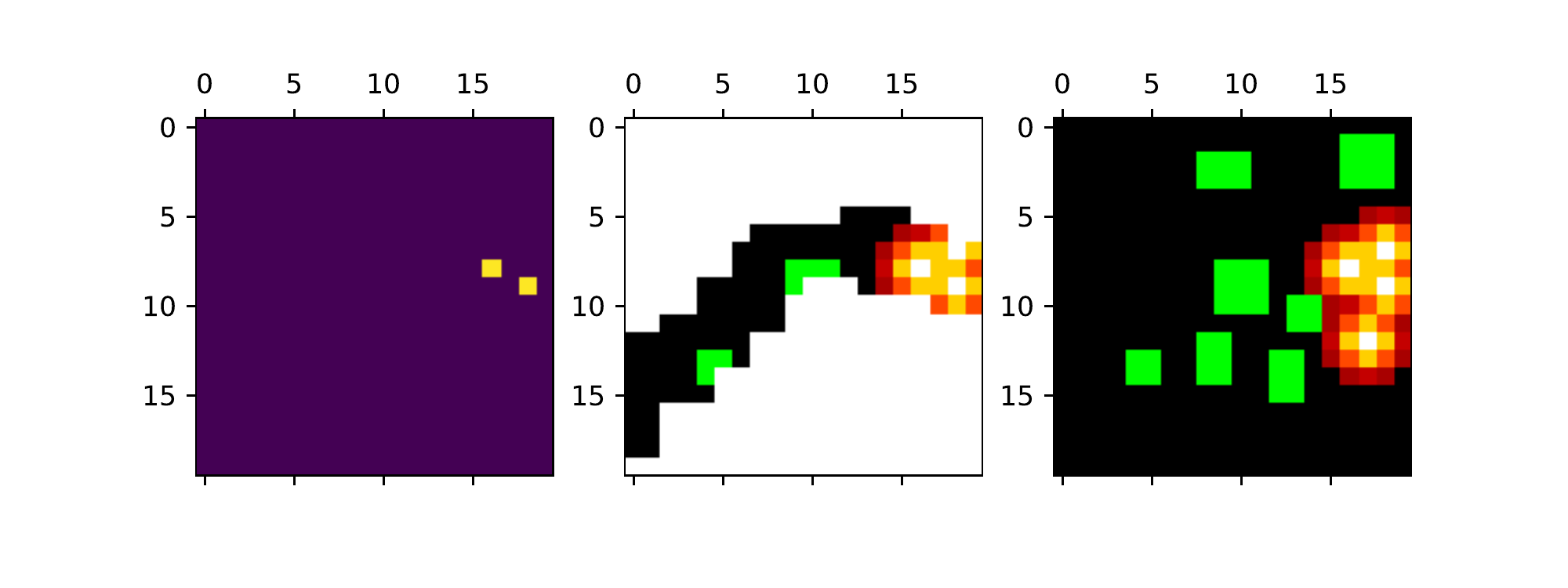}
        \label{fig:episode_end_obs}
    \end{subfigure}
        \vspace{-0.5cm}
    \caption{Drone positions (left), known map (center), real map (right). Beginning (above) and end (below) of an episode with obstacles.}
    \label{fig:episode_state_obs}
\end{figure}

\begin{figure}[t]
    \centering
    \begin{subfigure}{0.49\textwidth}
        \centering
\includegraphics[width=\textwidth,trim=0cm 0.5cm 1cm 2cm,clip]{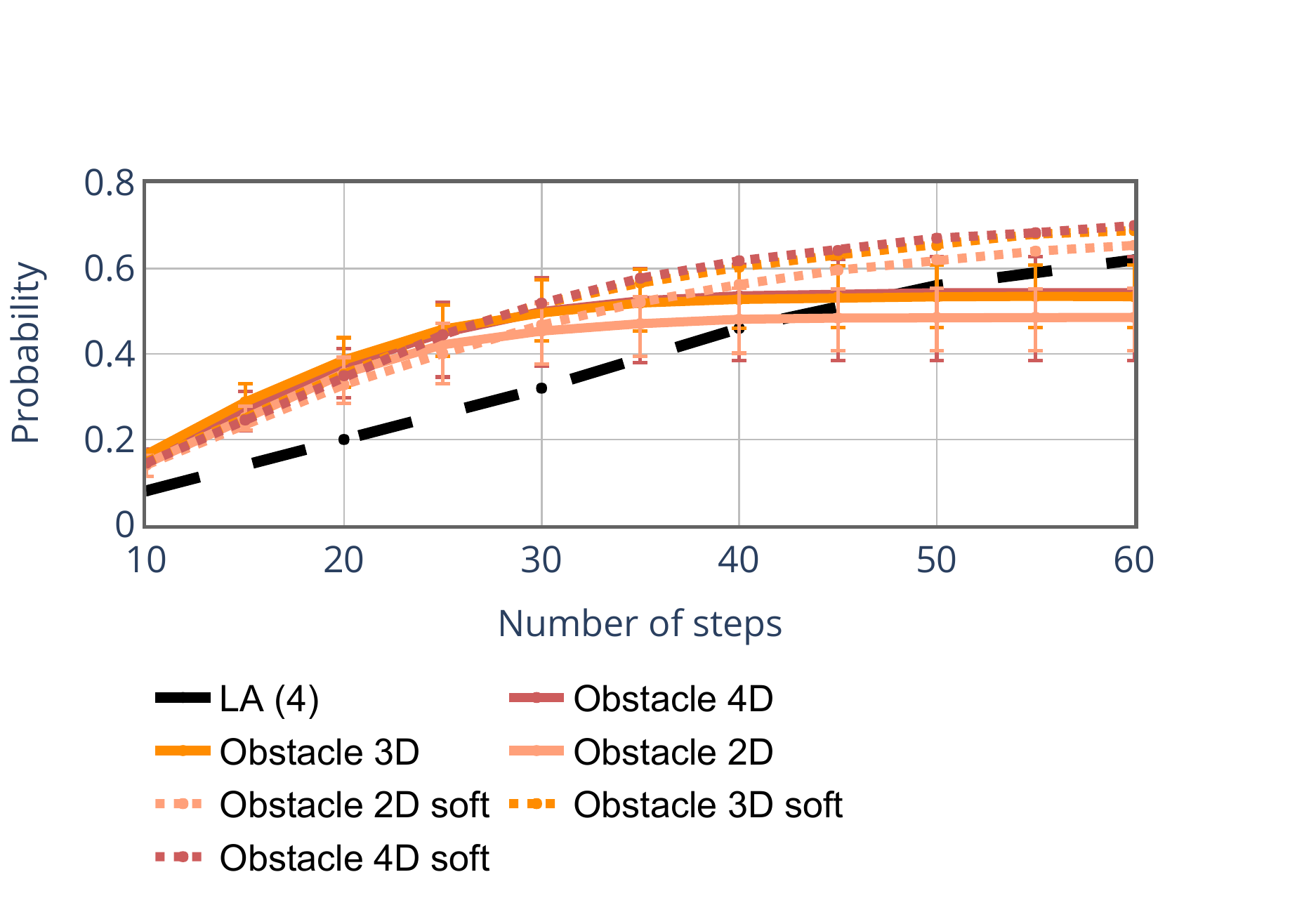}
\caption{2 \glspl{uav}.}
\label{fig:Obstacle}
    \end{subfigure}
    \begin{subfigure}{0.49\textwidth}
        \centering
\includegraphics[width=\textwidth,trim=0cm 0.5cm 1cm 2cm,clip]{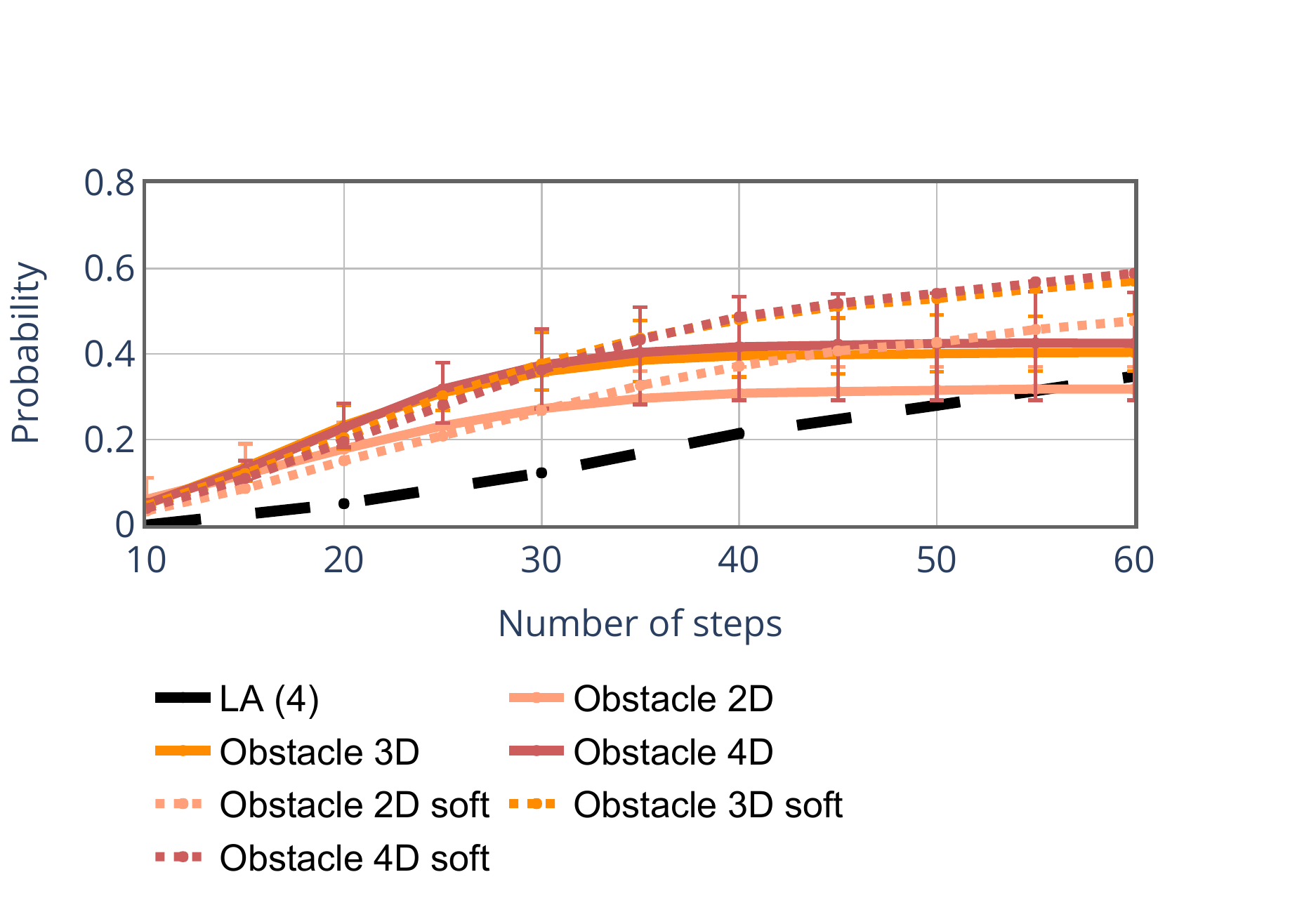}
\caption{3 \glspl{uav}.}
\label{fig:Obstacle_3d}
    \end{subfigure}
    \caption{CDF of the episode duration for different algorithms in the obstacle scenario.}
    \label{fig:obs}
\end{figure}

\edit{In what follows, we consider a modified version of the cluster scenario, where some obstacles are added to the map.
In particular, we empirically set the percentage of the map occupied by obstacles to 10\%, searching for a balance between increased system complexity and the realism of the scenario. An example of the system state representation with obstacles is shown in Fig.~\ref{fig:episode_state_obs} at the beginning and at the end of an episode. The obstacles are marked in green.}

Fig.~\ref{fig:Obstacle} shows the performance of the LA approach and \gls{ddql} in the case of 2 \glspl{uav} \edit{and 4 targets}. The \gls{ddql} solution has been trained for scenarios with 2, 3 and 4 \glspl{uav} \edit{(labeled in the plots as 2D, 3D, and 4D, respectively)}, and then tested in the scenario with 2 and 3 \glspl{uav}, with and without the use of the softmax approach in the testing phase. In both cases, it is clear that the models trained with more \glspl{uav} are able to outperform those with fewer \glspl{uav} in both considered scenarios. Furthermore, as for the case without obstacles, the use of the softmax policy during the testing phase increases the performance, especially when the episodes are longer, as it keeps the \glspl{uav} from getting stuck. In the scenario with 3 \glspl{uav}, as in Fig.~\ref{fig:Obstacle_3d}, the performance is generally lower, meaning that the swarm needs more training. \edit{However, \gls{ddql} is able to outperform the LA approach in both cases, reaching targets significantly faster in the scenario with 3 drones.}

\subsection{Transfer learning on bigger maps, with larger swarms \edit{and communication impairments}}

\begin{figure}[t]
    \centering
    \begin{subfigure}{0.49\textwidth}
        \centering
\includegraphics[width=\textwidth]{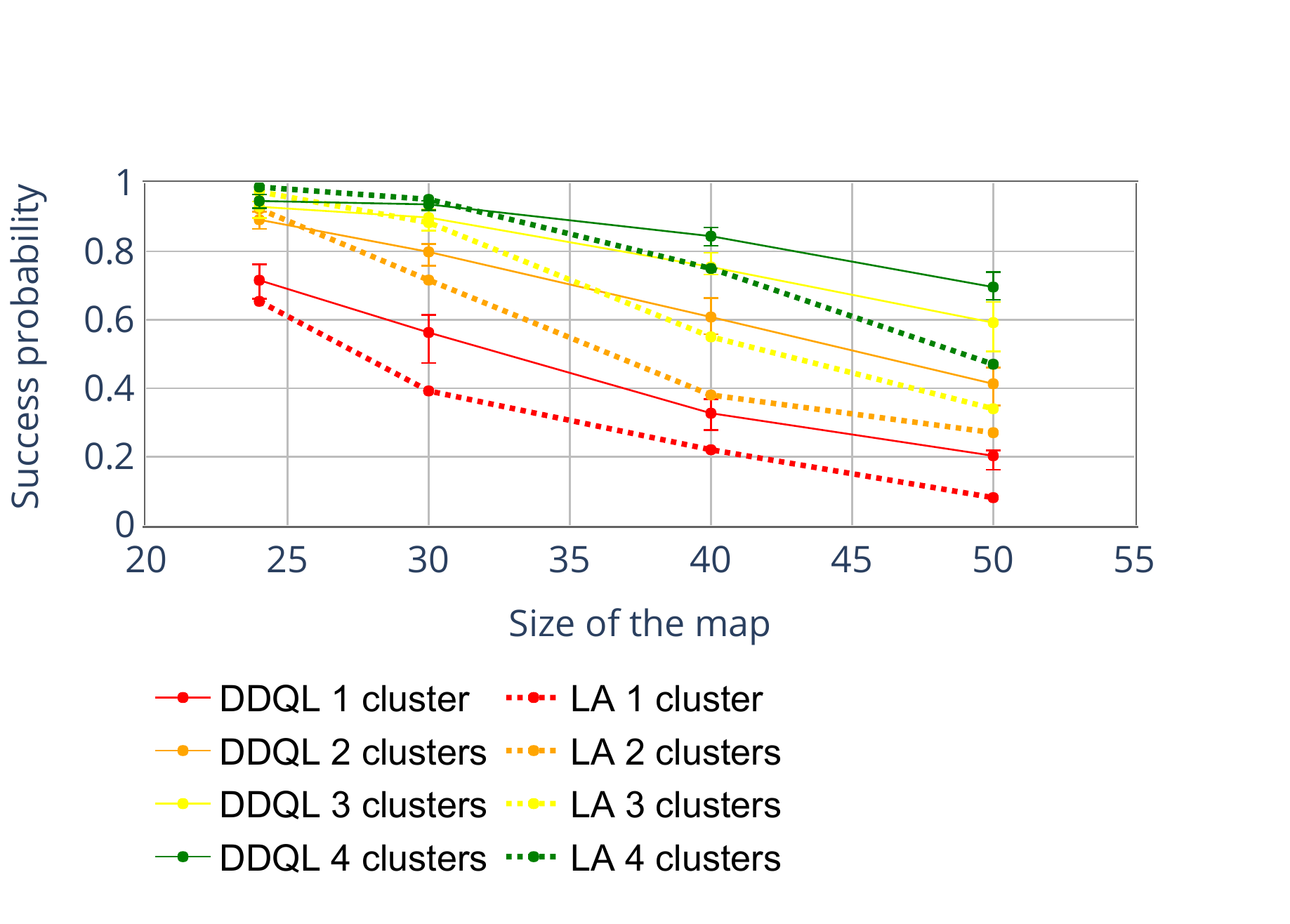}
\caption{2 \glspl{uav}.}
\label{fig:Bigmap2drones}
    \end{subfigure}
    \begin{subfigure}{0.49\textwidth}
        \centering
\includegraphics[width=\textwidth]{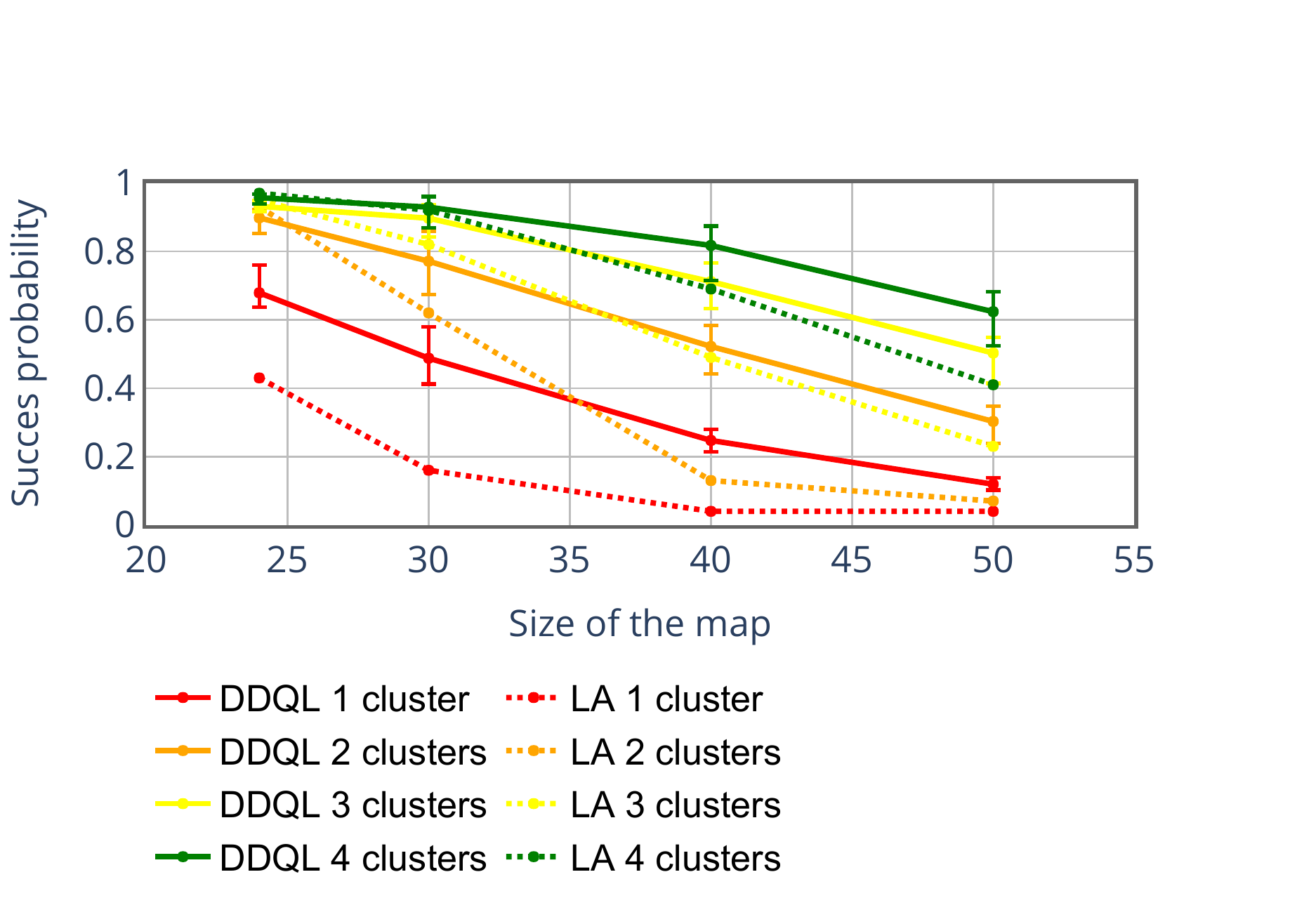}
\caption{3 \glspl{uav}.}
\label{fig:Bigmap3drones}
    \end{subfigure}
    \caption{Success probability as a function of the map size and the number of clusters.}
    \label{fig:final_sit}
\end{figure}

\begin{figure}[t]
    \centering
    \begin{subfigure}{0.49\textwidth}
        \centering
\includegraphics[width=\textwidth]{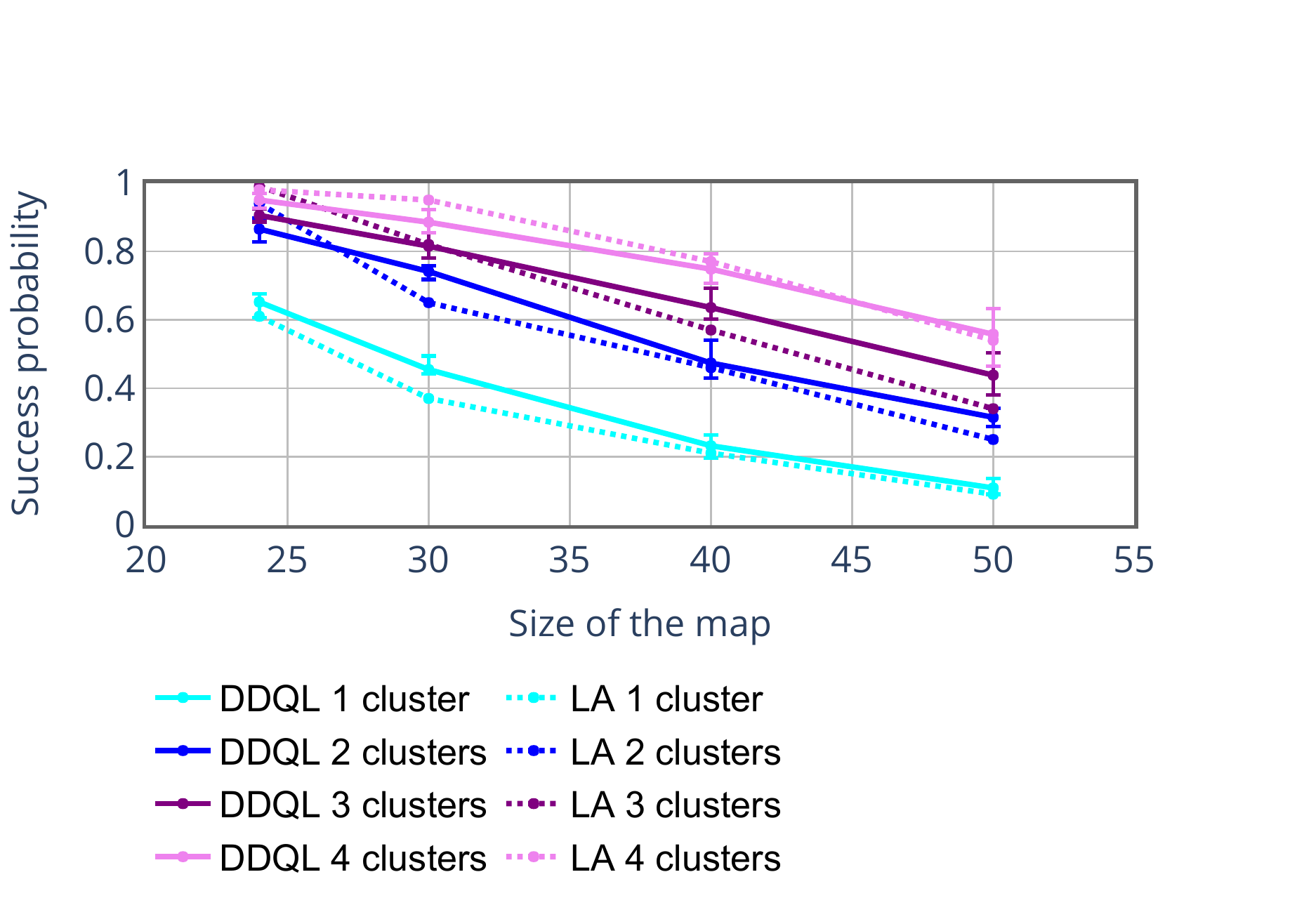}
\caption{2 \glspl{uav}.}
\label{fig:Bigmap2dronesobstacle}
    \end{subfigure}
    \begin{subfigure}{0.49\textwidth}
        \centering
\includegraphics[width=\textwidth]{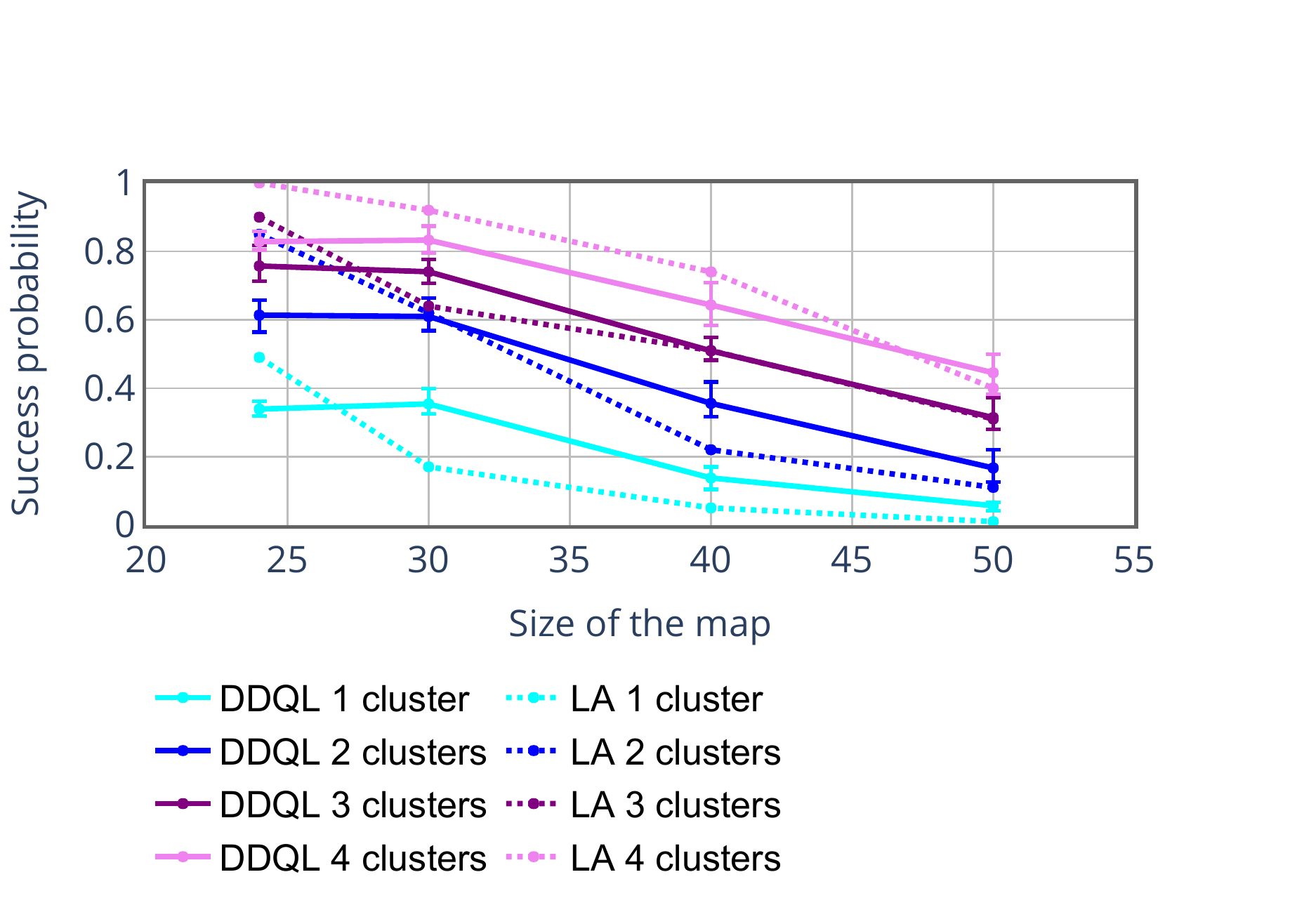}
\caption{3 \glspl{uav}.}
\label{fig:Bigmap3dronesobstacle}
    \end{subfigure}
    \caption{Success probability as a function of the map size and the number of clusters with obstacles.}
    \label{fig:final_sit}
\end{figure}

We then show how well \gls{ddql} is able to generalize to bigger maps in the testing phase. For this reason, the algorithm has been trained on a map with $M=24$, maintaining $F=20$, and the testing phase included bigger maps and different numbers of clusters. All the results shown in the following figures are obtained with 100-step episodes: the longer duration is needed to allow the agents to reach the targets even in bigger maps. For similar reasons, the scenarios with more clusters are studied to maintain a similar proportion of surface occupied by targets even in the bigger maps. Fig.~\ref{fig:Bigmap2drones} and Fig.~\ref{fig:Bigmap3drones} show how the performance varies as a function of the size of the environment and the number of clusters present in the map. In both cases, \gls{ddql} shows a good adaptability, getting better performance than LA in all cases, with a bigger gain in bigger maps. In Fig.~\ref{fig:Bigmap2dronesobstacle} and Fig.~\ref{fig:Bigmap3dronesobstacle}, the same scenarios are studied with the addition of the obstacles in the map, covering about 10\% of the size of the map. In this case, \gls{ddql} \edit{will need some retraining to reach LA's performance} on smaller maps, while the performance is similar when the map is bigger. However, we recall that \gls{ddql} also has a significant advantage in terms of computational cost, so it is preferable if performance is similar.

\begin{figure}[t]
    \centering
    \begin{subfigure}{0.49\textwidth}
        \centering
        \includegraphics[width=\linewidth]{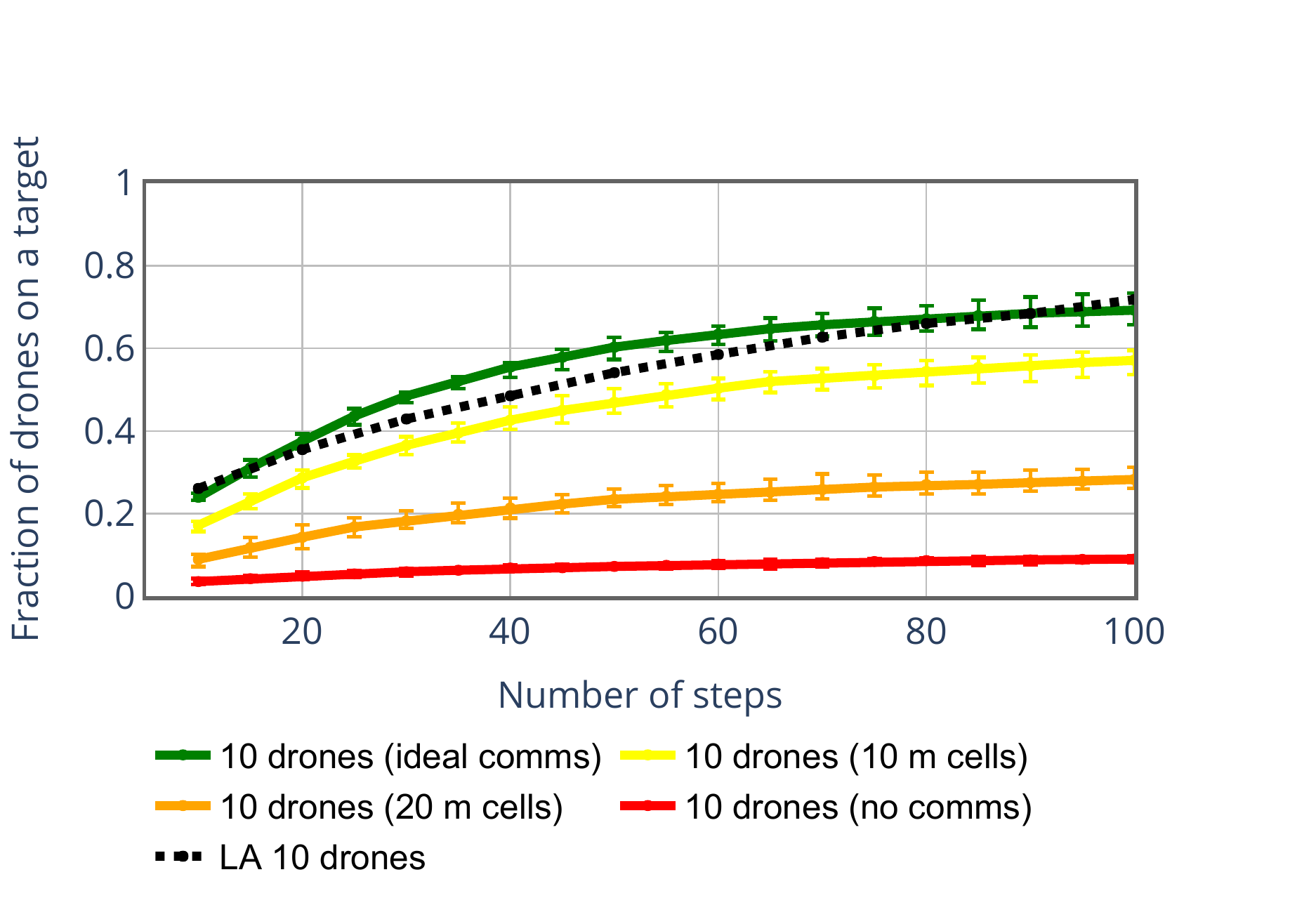}
        \caption{Performance of a swarm of 10 UAVs.}
        \label{fig:Dronescomm10}
    \end{subfigure}
    \begin{subfigure}{0.49\textwidth}
        \centering
\includegraphics[width=\linewidth]{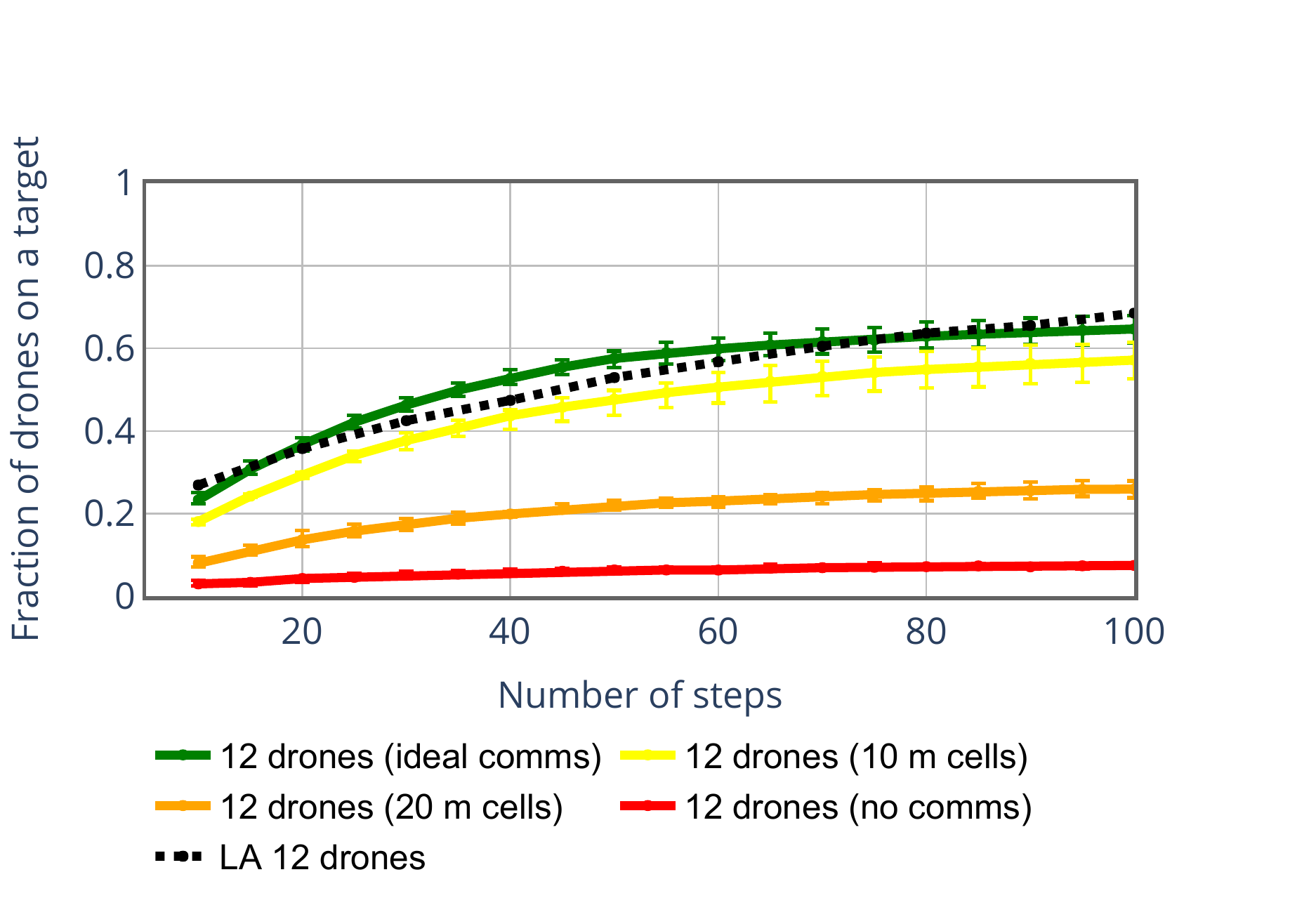}
\caption{Performance of a swarm of 12 UAVs.}
\label{fig:Dronescomm12}
    \end{subfigure}
    \caption{Effect of imperfect communications on the performance of DDQL in a large map.}
    \label{fig:comms}
\end{figure}

\edit{It is also interesting to test the transfer capabilities of the algorithms in more complex scenarios, including far larger swarms and imperfect communications: as \gls{ddql} relies on information from other \glspl{uav} to find targets and avoid collisions, a limited communication range can impair its performance significantly. As Fig.~\ref{fig:Dronescomm10} shows, 10 drones moving in a large map with obstacles (with 16 targets in 4 clusters, as above) can coordinate effectively with no retraining, outperforming the LA approach. Performance loss is limited even with communication restrictions if each cell is a square with a 10 m side, corresponding to a maximum range of about 11 cells with 50\% packet loss at the boundary of the coverage area. Performance loss with respect to the perfect communication scenario is limited, confirming the intuitive idea that information from neighbors inside the visible area is the most critical to find and reach the targets. If the cell side is doubled, effectively halving the communication range and introducing significant errors even for packets between immediate neighbors, the performance drops significantly, and becomes even worse if there is no communication at all between the \glspl{uav}. This would be true for any cooperative algorithm, as information from other \edit{agents} can be used to optimize the exploration of the map, but we highlight that \gls{ddql} has always been trained assuming ideal communication, and the communication impairments have been considered only in the test phase. Therefore, the \glspl{uav} might be confused by the lack of information, and a partial retraining might yield better results as the agents transfer their experience and learn to deal with the more limited feedback. On the other hand, the algorithm scales extremely well to larger swarms, slightly outperforming \edit{LA} even with no retraining in the new scenario. The same pattern holds for the case with 12 drones, which is shown in Fig.~\ref{fig:Dronescomm12}.}

\begin{figure}[h!]
\centering
\begin{subfigure}{0.9\textwidth}
  \centering
  \includegraphics[width=1\linewidth]{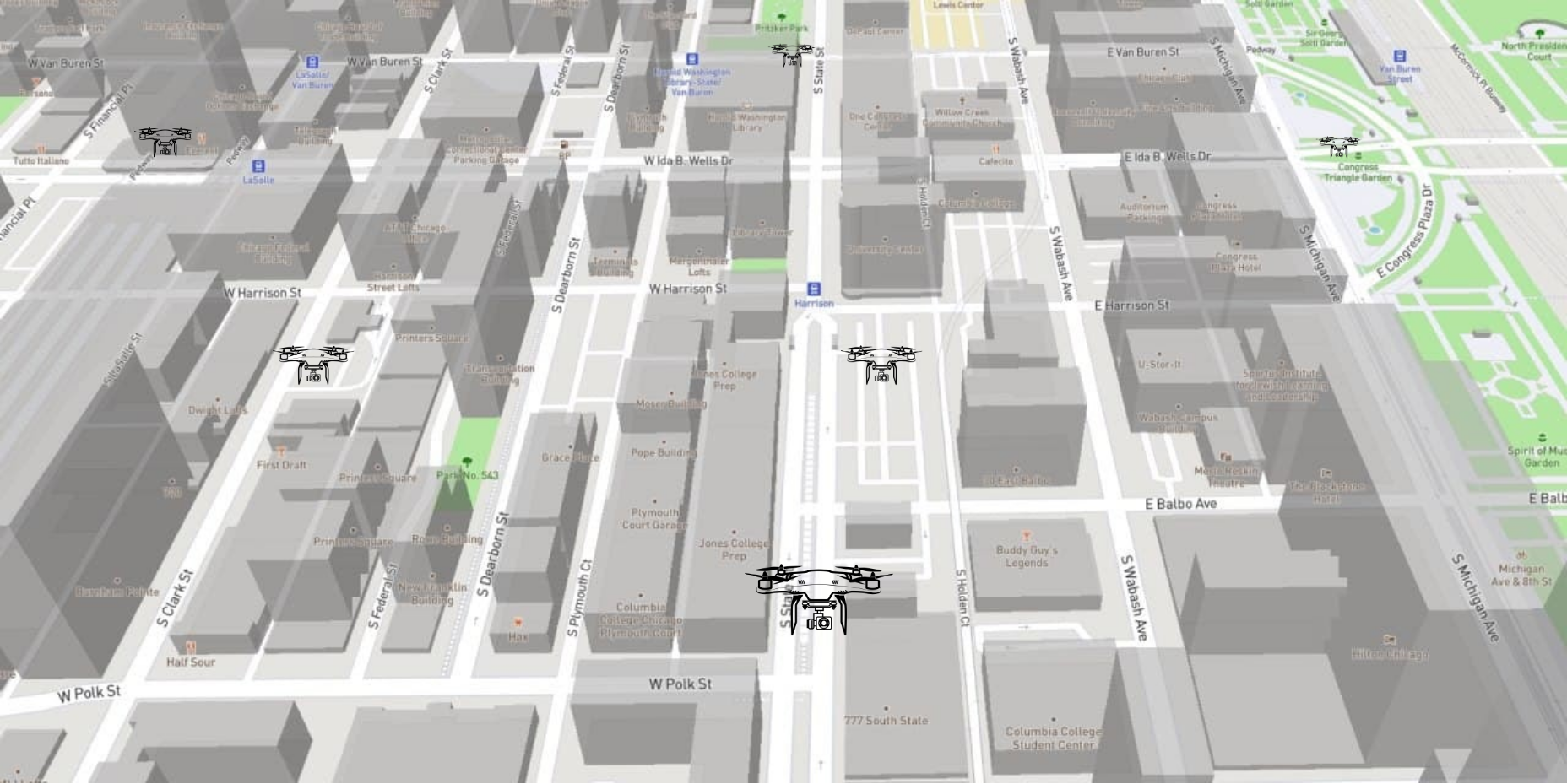}  
  \label{fig:sub-first}
\end{subfigure}
\begin{subfigure}{.29\textwidth}
  \centering
  \includegraphics[width=1\linewidth]{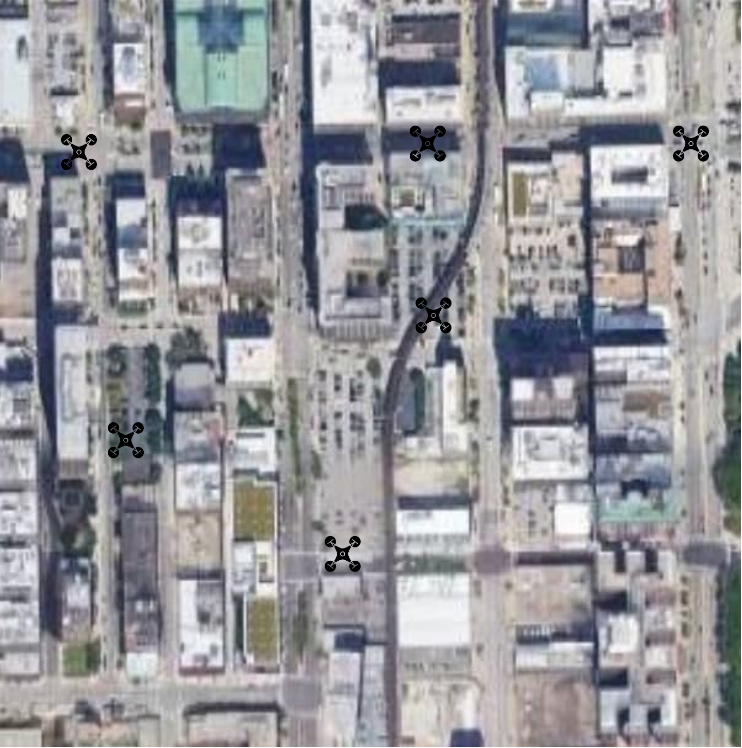}  
  \label{fig:sub-second}
\end{subfigure}
\begin{subfigure}{.29\textwidth}
  \centering
  \includegraphics[width=1\linewidth]{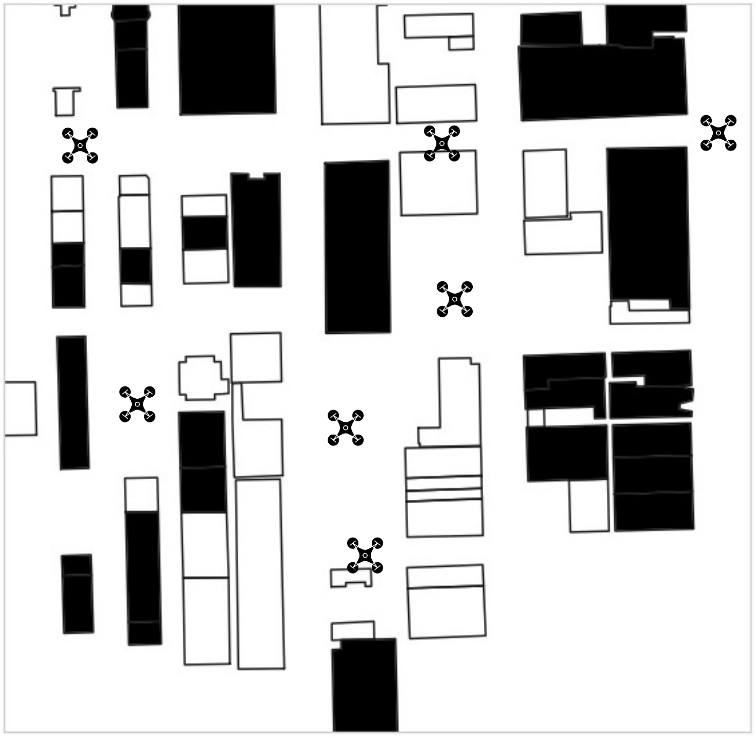}  
  \label{fig:sub-second}
\end{subfigure}
\begin{subfigure}{.29\textwidth}
  \centering
  \includegraphics[width=1\linewidth]{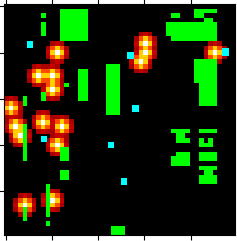}  
  \label{fig:sub-second}
\end{subfigure}
\caption{Extraction of the map from building height data in a 500 m by 500 m area in the downtown Chicago Loop neighborhood.}
\label{fig:chicago_maps}
\end{figure}

\begin{figure}[t!]
\centering
\includegraphics[width=.7\textwidth]{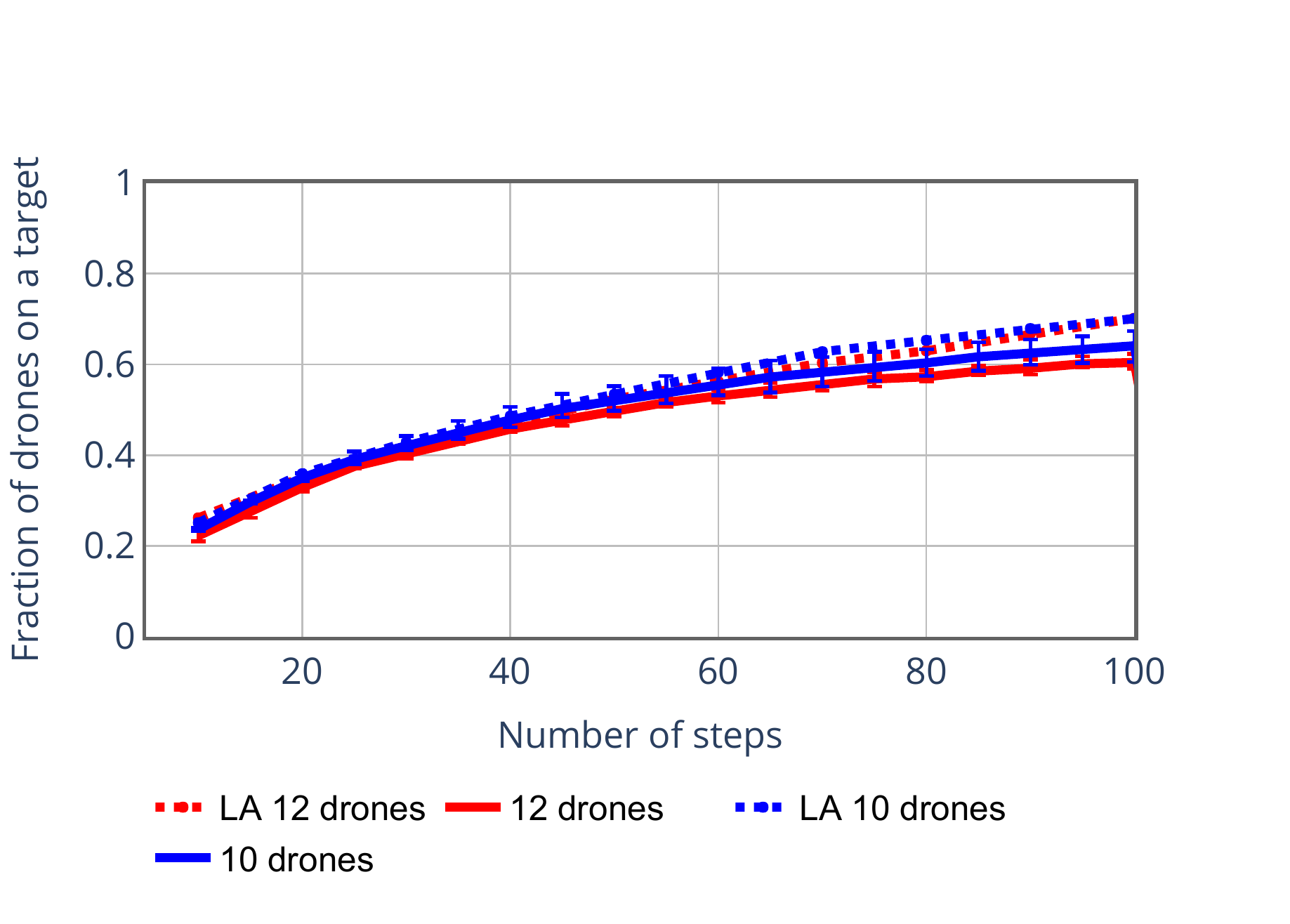}
\caption{Performances on the real map of Chicago}
\label{fig:Chicagoperf}
\end{figure}

\edit{Finally, we tested an extreme transfer learning scenario, not only increasing the size of the map and the number of \glspl{uav}, but also switching from the synthetic obstacle distribution on the map to one derived from a real map. The map of obstacles was obtained from a city map of the area just east of LaSalle Street Station in downtown Chicago, in the central Loop neighborhood. As shown in Fig.~\ref{fig:chicago_maps}, we obtained the height profiles of the buildings in the area, considering as obstacles their parts with a height of over 10 stories (i.e., approximately 40 m, the minimum legal hovering altitude for \glspl{uav}). The 500 m by 500 m area was then divided into 2500 square cells with 10 m sides, converting the height profile to an obstacle map in the grid. 11\% of the map was occupied by obstacles, so the map was approximately as full as the one used in the training, which had 10\% obstacle cover, but the individual obstacles were larger and concentrated along South Wabash Avenue and South Dearborn Street. This is an additional hurdle for \gls{ddql}, which was not trained to deal with obstacles concentrated along streets, which make it more difficult to find an appropriate path. However, as Fig.~\ref{fig:Chicagoperf} shows, the \gls{ddql} system can find targets approximately as fast as LA, underperforming a little only on higher percentiles. With a modicum of retraining, \gls{ddql} should be able to adapt to the different structure, exploiting the regularities in city blocks to avoid obstacles and find targets even more quickly.}

\edit{In conclusion, we have shown that \gls{ddql} is able to find efficient strategies for the \glspl{uav} to reach targets faster than look-ahead solutions in complex environments. The algorithm is scalable to larger maps, larger swarms, and limited communications without any retraining, and can deal with obstacles and very different target distributions with a limited amount of retraining. This shows that the solution is powerful and versatile, adapting easily to new conditions. However, there is still some margin for improvement, particularly in scenarios in which almost all the \glspl{uav} in the swarm have already reached a target, while the last stragglers are far from any feature in the map. This case represents most of the residual failures of the algorithm, and solving it is an important future objective.}

\section{Conclusions and future work} \label{sec:conclusions}

In this work, we \edit{studied} the problem of area monitoring and surveillance with a swarm of drones. We modeled the environment with a 2D grid and cast the problem into the theoretical framework of \gls{ndpomdp}. We have examined various scenarios, including obstacles and maps of different sizes, and the \edit{proposed} algorithm outperformed a computationally intensive look-ahead approach in almost all scenarios.

Important research directions include the introduction of dynamic targets, which would be an important step to increase the scenario's realism, as well as different roles for the drones, which can be assigned dynamically and would allow us to examine another interesting aspect of the \gls{marl} problem, increasing the difficulty of coordinating the \glspl{uav}' actions.

\section*{Acknowledgments}
This work has been \edit{partially} supported by the U.S. Army Research Office (ARO) under Grant no. W911NF1910232, "Towards Intelligent Tactical Ad hoc Networks (TITAN)" and \edit{by} MIUR (Italian Ministry for Education and Research) under the initiative "Departments of Excellence" (Law 232/2016)".

\bibliographystyle{IEEEtran}
\bibliography{bibliography.bib}


\end{document}